\pdfoutput=1

\documentclass[11pt]{article}

\usepackage[]{acl}

\usepackage{times}
\usepackage{latexsym}

\usepackage[T1]{fontenc}

\usepackage[utf8]{inputenc}

\usepackage{microtype}

\usepackage{inconsolata}

\usepackage{amsmath}
\usepackage{graphicx}
\graphicspath{{figures/survey}}
\usepackage[]{todonotes}
\usepackage{booktabs}
\usepackage{makecell}
\usepackage{enumitem}
\usepackage{url}
\usepackage{fontawesome}
\usepackage{pifont}
\usepackage{array}
\usepackage{multirow}
\usepackage{xcolor}
\usepackage{colortbl}
\usepackage{subcaption}
\usepackage[belowskip=-0.2em,labelfont=bf,labelsep=endash]{caption}
\usepackage[round-mode=places,round-precision=0,group-separator={,},output-decimal-marker={.}]{siunitx}
\usepackage[nolist]{acronym}

\begin{acronym}
    \acro{wals}[WALS]{the World Atlas of Language Structures}
    \acro{nlp}[NLP]{natural language processing}
\end{acronym}

\newcommand{\eq}[0]{\textsuperscript{*}}
\newcommand{\kul}[0]{\textsuperscript{\ding{168}}}
\newcommand{\aal}[0]{\textsuperscript{\ding{95}}}

\newcommand{\acctext}[0]{\ding{169}}
\newcommand{\maptext}[0]{\ding{72}}
\newcommand{\ftext}[0]{\ding{171}}

\newcommand{\acc}[0]{\textsuperscript{\acctext}}
\newcommand{\map}[0]{\textsuperscript{\maptext}}
\newcommand{\f}[0]{\textsuperscript{\ftext}}

\newcommand{\nolangtext}[0]{\dag}
\newcommand{\missingtext}[0]{\ddag}
\newcommand{\lowcovtext}[0]{\textsection}

\newcommand{\nolang}[0]{\textsuperscript{\nolangtext}}
\newcommand{\missing}[0]{\textsuperscript{\missingtext}}
\newcommand{\lowcov}[0]{\textsuperscript{\lowcovtext}}

\newcommand\nonumberfootnote[1]{%
  \begingroup
  \renewcommand\thefootnote{}\footnote{#1}%
  \addtocounter{footnote}{-1}%
  \endgroup
}

\newcommand*{\MinNumber}{-15.0}%
\newcommand*{\MidNumber}{0.0} %
\newcommand*{\MaxNumber}{5.0}%

\definecolor{high}{HTML}{03AC13}
\definecolor{mid}{HTML}{F7E379}
\definecolor{low}{HTML}{ec462e}
\newcommand*{\opacity}{80}

\newcommand{\tgrad}[1]{%
    \ifdim #1 pt > \MidNumber pt%
        \pgfmathparse{max(min(100.0*(#1 - \MidNumber)/(\MaxNumber-\MidNumber),100.0),0)}%
        \xdef\PercentColor{\pgfmathresult}%
        \cellcolor{high!\PercentColor!mid!\opacity}#1%
    \else
        \pgfmathparse{max(min(100.0*(\MidNumber - #1)/(\MidNumber-\MinNumber),100.0),0)}%
        \xdef\PercentColor{\pgfmathresult}%
        \cellcolor{low!\PercentColor!mid!\opacity}#1%
    \fi
}

\definecolor{low-color}{HTML}{621F87}
\definecolor{high-color}{HTML}{D16232}

\title{What is ``Typological Diversity'' in NLP?}

 \author{Esther Ploeger\eq\aal \quad 
  Wessel Poelman\eq\kul \quad 
  Miryam de Lhoneux\kul \quad 
  Johannes Bjerva\aal\\
  \aal Department of Computer Science, Aalborg University, Denmark \\ 
  \kul Department of Computer Science, KU Leuven, Belgium \\ 
  \texttt{\{espl,jbjerva\}@cs.aau.dk} \quad
  \texttt{\{wessel.poelman,miryam.delhoneux\}@kuleuven.be}}

\begin{document}
\maketitle

\nonumberfootnote{* Equal contribution.}

\begin{abstract}
The NLP research community has devoted increased attention to languages beyond English, resulting in considerable improvements for multilingual NLP.
However, these improvements only apply to a small subset of the world's languages.
An increasing number of papers aspires to enhance \textit{generalizable} multilingual performance \textit{across languages}.
To this end, linguistic typology is commonly used to motivate language selection, on the basis that a broad typological sample ought to imply generalization across a broad range of languages. 
These selections are often described as being \mbox{`typologically~diverse'}.
In this meta-analysis, we systematically investigate NLP research that includes claims regarding \mbox{typological~diversity}.
We find there are no set definitions or criteria for such claims.
We introduce metrics to approximate the diversity of resulting language samples along several axes and find that the results vary considerably across papers.
Crucially, we show that skewed language selection can lead to overestimated multilingual performance.
We recommend future work to include an operationalization of typological diversity that empirically justifies the diversity of language samples. To help facilitate this, we release the code for our diversity measures.\footnote{Our code and data are publicly available: \href{https://github.com/WPoelman/typ-div-survey}{\texttt{https://github.com/WPoelman/typ-div}}}
\end{abstract}

\section{Introduction}

Most research in the field of \ac{nlp} is conducted on the English language \citep{ruder-etal-2022-square}.
Competitive monolingual language modelling beyond English remains challenging, as current state-of-the-art methods rely on the availability of large amounts of data, which are not available for most other languages \citep{joshi-etal-2020-state}.
\begin{figure}
    \centering
    \includegraphics[width=\columnwidth]{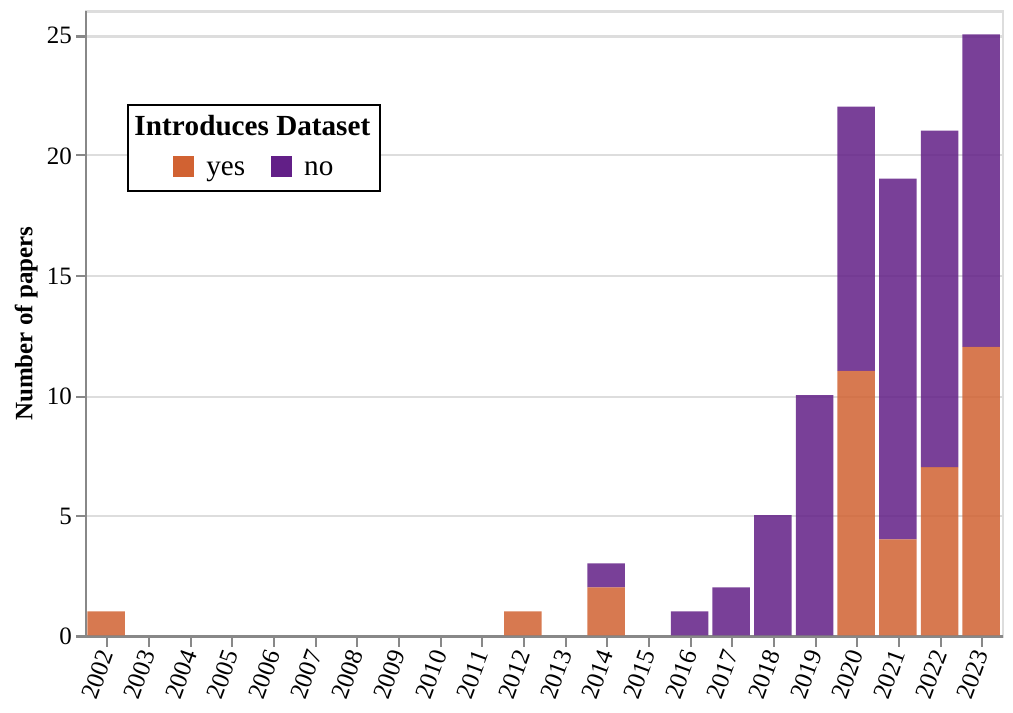}
    \caption{There is an increase in the number of publications with `typological diversity' claims across time.}
    \label{fig:number-of-papers}
\end{figure}
This data sparsity can be mitigated by leveraging cross-lingual transfer through the training of a language model on multilingual data. %
Despite the potential of multilingual language modelling, methodologies are primarily developed for English. %
However, there is no guarantee that an approach that works well for one language will work equally well for others \citep{gerz-etal-2018-relation}. 
For instance, morphologically complex (or rich) languages can be over-segmented by current widely-used tokenization methods \citep{rust-etal-2021-good}. 
Evaluation on a broad range of languages is important for drawing more generalizable conclusions about the performance of multilingual language technology \cite{bender-2009-linguistically,bender2011achieving,pikuliak-simko-2022-average}.
Evaluating a tokenization method on only morphologically simple languages, such as English, can give an unrealistic image of the effectiveness of a tokenization method, because morphologically simple languages are generally easier to tokenize compared to complex ones.

Current work increasingly evaluates models on multiple languages, but because of resource constraints, it is not realistic to test a model on the thousands of languages in the world.
In order to still ensure a degree of generalizability, previous work recognizes the importance of diverse language sampling.
\citet{ponti-etal-2020-xcopa} suggest that merely evaluating on a small set of similar languages is an unreliable method for estimating a multilingual model's performance, since such evaluations lack robustness to previously unseen typological properties, for instance. 
In an attempt to address 
these sampling issues, \ac{nlp} research increasingly claims to rely on linguistic typology (see Figure~\ref{fig:number-of-papers}). 
Evaluating on a set of so-called `typologically diverse' languages implies robustness over typological variations, implicitly claiming broad generalizability across languages. %
Importantly, however, there is a fundamental lack of common understanding of exactly what a claim of typological diversity entails.

In this work, we systematically investigate how the term `typological diversity' is used in \ac{nlp} research by surveying the literature for past and current practices. 
We find that there is no set definition of `typological diversity' in \ac{nlp}: justifications for claims range from (i) \textit{none whatsoever}, to (ii) using language phylogeny as a proxy for typology, to (iii) basing diversity on a more or less well-founded subset of available typological features.
As a consequence, typological language similarity and typological feature value inclusion vary considerably across papers (\S \ref{sec:analysis}).
Furthermore, we demonstrate why this is problematic, by showing how a skewed selection of evaluation languages can lead to overstated multilingual performance and poor generalizability (\S \ref{sec:model_eval}).
We recommend future work using the term `typologically diverse' to document their language samples, and to add a measure of typological diversity, such as the mean pairwise language distance (\S \ref{sec:dist}) or typological feature inclusion (\S \ref{sec:cov}).
Better-documented language sampling is crucial for the development of multilingual \ac{nlp}, because it provides  insights into commonly overlooked linguistic difficulties of multilingual language modelling.
We provide software to verify and visualize this, linked below the abstract.

\section{Background}
\label{sec:background}

Language sampling techniques are used in linguistic typology as well as in \ac{nlp}.
Although these fields are both concerned with languages and generalizability, they typically have different objectives and resources, which we contrast in this section.

\paragraph{Objectives}
Language sampling has a long-standing history in linguistic typology.
\citet{rijkhoff1993method} state that testing a linguistic hypothesis on a few similar languages is not sufficient for drawing  conclusions on human language as a whole.
Rather, they argue that generalizable samples should capture the diversity of the world's languages.
Language samples can be constructed in a number of ways. \citet{rijkhoff1998language} distinguish three types of sampling:
\textit{random}, \textit{probability} and \textit{variety}.
Random samples do not have any specific stratification criteria, %
whereas probability samples are constructed to minimize bias, for instance by avoiding a skew towards languages from particular language families or geographical areas. 
Variety samples are designed to contain as much linguistic variety as possible, including relatively uncommon typological characteristics \citep{miestamo2016}.
Finally, in addition to the three methods outlined, sampling can be based on data availability, which is defined as \textit{convenience} sampling \cite[p. 50]{velupillai2012introduction}.

Language sampling methods in \ac{nlp} have a similar goal: to sample languages in such a way that the findings (e.g., multilingual performance estimates) are expected to generalize across languages.
Yet, their justifications are commonly less rigorous than those in linguistic typology.
For language sampling methods in \ac{nlp}, it is usually not explicitly mentioned which type of sampling is conducted, and for which reasons.
In practice, most papers do not provide a justification for their selections (see \S \ref{sec:just}).
This is likely caused by data availability, albeit implicitly, which resembles convenience sampling.
A rare exception to this is \citet{ponti-etal-2020-xcopa}, who use variety sampling and mention that this ensures evaluation that is robust towards typological characteristics that were likely not seen during training.

\paragraph{Resources}
While the goal of language sampling for ensuring generalizable conclusions is similar between \ac{nlp} and linguistic typology, their resources are typically different.
In linguistic typology, the variables under investigation are often typological characteristics. 
A typologist may aim to test whether a correlation of features holds across languages. 
Such a correlation could then be referred to as a `universal': languages with object-verb ordering would be postpositional \cite{greenberg1963universals}, for instance.
To avoid circularity, language sampling is not done directly based on those features. 
Instead, methods commonly rely on (combinations of) language family, genus, and geography \citep{miestamo2016}.

By contrast, what is evaluated in multilingual \ac{nlp} does not concern typological features directly, rather, model performance across languages.
In fact, to support claims of generalizability, multilingual benchmarks must explicitly contain a range of typological phenomena \citep{bender-2009-linguistically,bender2011achieving,pikuliak-simko-2022-average}. %
Relying on geographical and genetic relations between languages is not guaranteed to ensure this. 
\citet{stoll2013capturing} argue that \textit{``the kind of variables that define genealogical groups and tree shapes have a very different nature from the kind of variables that define typological diversity''}, based on work from \citet{nichols1996comparative}.
For estimating the generalizability of a language sample in \ac{nlp} evaluation, it is possible, and even desired, to have direct insight into the actual typological properties that are included in the selection (see also \S \ref{sec:model_eval}).

This motivation is similarly voiced by \citet{samardzic-etal-2024-measure}, who introduce a data-driven method of relating language samples in particular datasets to a reference language sample that is assumed to be typologically diverse.
Yet, our work differs in crucial ways.
Our aims are to look at what it means to make \emph{claims} of typological diversity in current research practices and how to characterize the language samples such claims are made about.
Our methods work as standalone measures; we do not need a reference sample to compare against.
In addition, our approach is centered around languages, thus only requiring language descriptions, not entire datasets.
\citet{samardzic-etal-2024-measure} approximate `aggregated morphological features' through the average word length in a corpus, but this number may vary across domains within the same language.

\section{Systematic Review}
\label{sec:methodology}
We systematically investigate the use of the term `typological diversity' in \ac{nlp} research, by conducting a survey across relevant published work.

\subsection{Data Collection}
We retrieve papers from the entire ACL Anthology (*ACL, COLING, LREC, etc.) and from several top-tier AI venues where \ac{nlp} research is published (NeurIPS, ICLR, AAAI, etc.).\footnote{See Appendix \ref{sec:app-paper-details} for details.}
We select papers that contain the following search string in either their title or abstract:

\begin{quote}
\colorbox[RGB]{239,240,241}%
{\parbox{0.82\columnwidth}{\texttt{typolog.+?div.+?|div.+?typolog.+?}}}
\end{quote}

\noindent
Our search string aims to capture a broad range of formulations, including: ``typologically and genetically diverse languages'' \cite{xu-etal-2020-modeling}, ``languages of diverse typologies'' \cite{eskander-etal-2022-unsupervised} and ``diverse languages in terms of language family and morphological typology'' \cite{eskander-etal-2020-unsupervised}.
Another commonly used phrase is ``we evaluate on $N$ typologically \textit{different} languages''.
While we acknowledge that some researchers might (unintentionally) use the two claims interchangeably, we only focus on claims regarding `diversity'.
It is reasonable to state that languages are `typologically different', even though they might be similar (e.g., Dutch and German).

\subsection{Annotation Guidelines}
The first two authors annotated the retrieved papers according to the following criteria:

\begin{enumerate}[noitemsep,topsep=5pt,parsep=5pt,partopsep=0pt]
    \item Does the paper claim that a particular language set is typologically diverse?
\end{enumerate}
If so, we annotate the following:
\begin{enumerate}[noitemsep,topsep=5pt,parsep=5pt,partopsep=0pt]
\setcounter{enumi}{1}
    \item Does the paper introduce a dataset?
    \item Which languages are included?
\end{enumerate}

For criterion 1, we only consider claims made by the paper at hand. 
That is, if a paper uses a dataset which is claimed to be `typologically diverse', but the current paper does not mention this, we do not consider this to be a claim.
This also applies to related work: if a paper references a claim made by another paper, but they themselves do not make a claim, then we do not consider this as a claim.
The reason that we annotate whether a paper introduces a new dataset is that datasets are often starting points for subsequent projects.
If datasets are claimed to be `typologically diverse', this claim is spread not only through the paper introducing it, but also the dataset itself.\footnote{An overview of all papers introducing a dataset for \mbox{`typologically~diverse'} languages is in Appendix~\ref{app:datasets}, Table~\ref{tab:dataset}.}

Lastly, we annotate which languages are included in the papers making a claim.
These are normalized to ISO-639-3 codes if available.
Where applicable, we only select the languages about which the claim is made.
For example, \citet{kann2020weakly} and \citet{shi2023language} specifically distinguish between baseline languages and their typologically diverse test language selection.

\subsection{Inter-Annotator Agreement}
In order to determine the agreement between the annotators, and estimate the quality of the annotation guidelines, we calculate Cohen's $\kappa$, as there are two annotators and we treat the values as binary.
The agreement values of all annotation items are shown in Table~\ref{tab:annotations}.
\begin{table}[h]
    \centering
    \small
    \begin{tabular}{lll}
    \toprule
        \textbf{Annotation item} & $\kappa$ & \textbf{Agreement} \\
        \midrule
         1. \texttt{has\_claim} & 0.78 & Substantial \\
         2. \texttt{introduces\_dataset} & 0.71 & Substantial \\
         3. \texttt{iso\_codes} & 0.44 & Moderate \\
         \bottomrule
    \end{tabular}
    \caption{Inter-annotator agreement per item.}
    \label{tab:annotations}
\end{table}

For the annotations concerning claims (1) and datasets (2), we achieve substantial agreement.
We resolve the disagreements by discussing the annotations and merging them into a single label.
For the sake of transparency, and to support incentives for human label variation \cite{plank-2022-problem}, we also release the non-aggregated annotations.\footnote{At the repository listed below the abstract.}
The agreement for the language selection annotation (3) is somewhat low, which is partially explained by the fact that we calculate agreement over all languages together, rather than on the micro-level for individual languages. This is because there is no ground truth regarding the number of languages to annotate.
Furthermore, a number of inconsistencies are due to ISO-639-3 variants (\texttt{jap}~$\rightarrow$~\texttt{jpn} or \texttt{ger}~$\rightarrow$~\texttt{deu}) and different codes for possibly ambiguous language mentions (`Norwegian' could be \texttt{nor} or \texttt{nob}).
We iteratively resolve these issues to get our final list of languages per paper.

\begin{figure}[hb]
    \centering
    \includegraphics[width=\columnwidth]{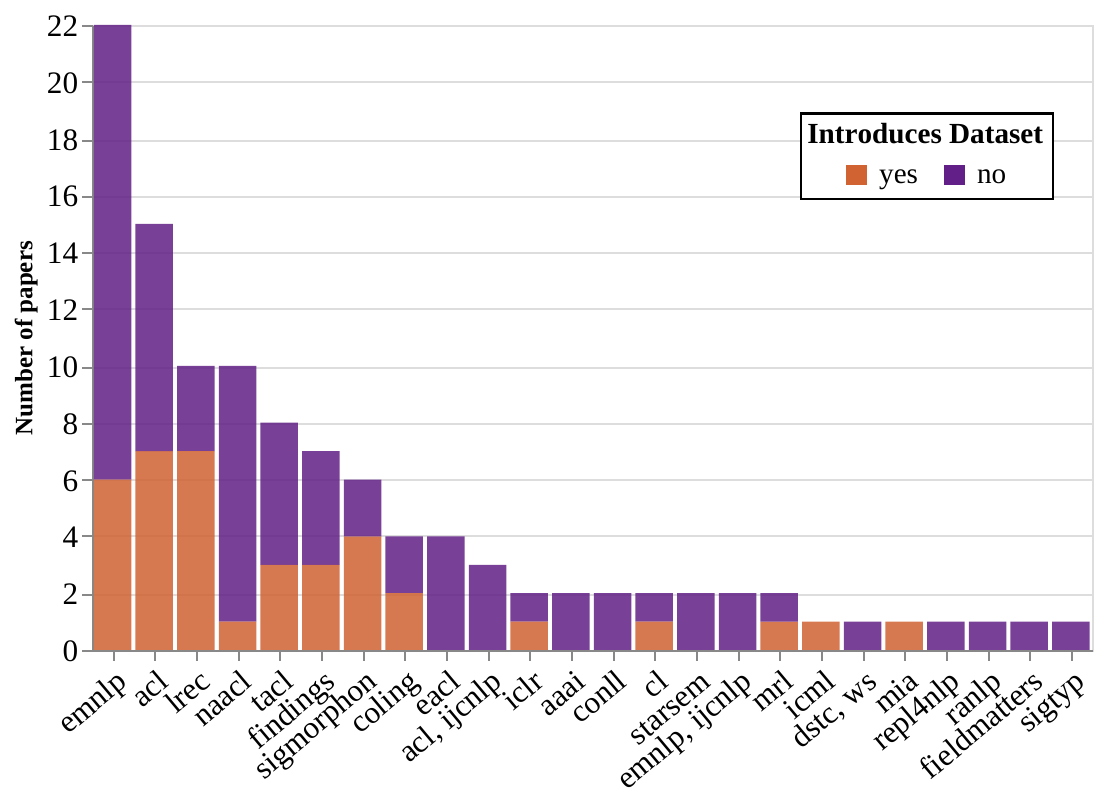}
    \caption{Number of papers with a claim by venue.}
    \label{fig:venues}
\end{figure}

\section{Results}
\label{sec:results}
\subsection{Publication Overview}
In total we retrieve 194 papers, of which 110 are found to contain a claim.
The rest of this analysis is based on these 110 papers.
Figure~\ref{fig:venues} shows the most common venues where papers with a claim have been published.
The top six venues are all high-ranking \ac{nlp} conferences, with SIGMORPHON being the first workshop at the seventh spot.
The AI-centered venues only have a small number of papers with claims; ICLR and AAAI both have two, ICML one, and the rest none.
In total, 38 papers introduce a dataset, most of which are published at ACL and LREC since 2020 (Figure~\ref{fig:number-of-papers} and \ref{fig:venues}).\footnote{Details are listed in Appendix \ref{app:datasets}, Table~\ref{tab:typ-datasets}.}

\subsection{Language Overview}
\label{sec:results-languages}

\begin{figure}[hb]
    \centering
    \includegraphics[width=\columnwidth]{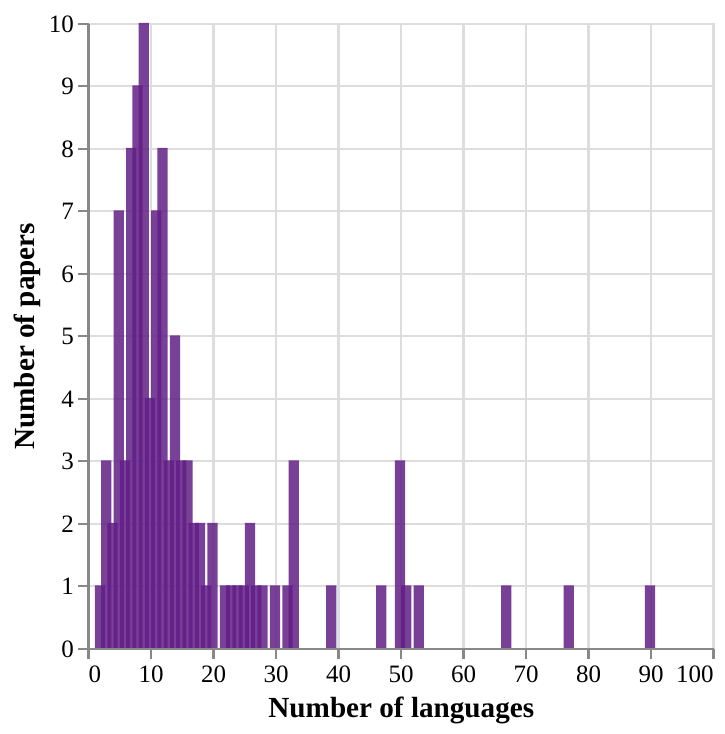}
    \caption{Papers by number of used languages.}
    \label{fig:num_langs}
\end{figure}

\begin{figure*}[t]
   \centering
\includegraphics[width=\textwidth]{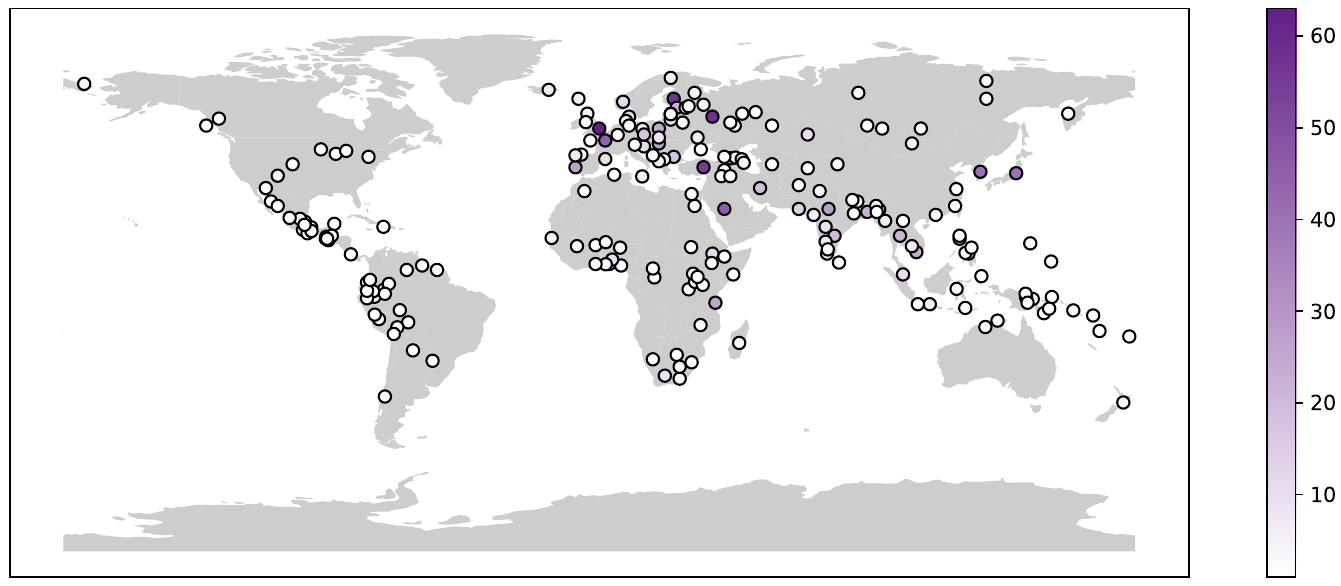}
    \caption{Map of languages in all papers claiming `typological diversity', where the hue corresponds number of papers that uses a language. Coordinates are taken from WALS.}
    \label{fig:map}
\end{figure*}

Four of the papers that claim typological diversity do not mention which languages they use at all.
The other papers use a median of 11 languages ($Q1\!=\!8$, $Q3\!=\!18$), with a minimum of 2 and a maximum of 90. %
In total, the papers use 315 unique languages, of which 160 are used just once. 
Figure~\ref{fig:num_langs} shows the distribution of the number of languages used per paper. 

English is the most used language (63 papers), followed by German (60), Russian (58) and Finnish (57).\footnote{The long tail of this distribution is partially included in Appendix \ref{sec:app-long-tail}, Figure~\ref{fig:long_tail}.} 
The paper that uses the most languages is \citet{vylomova-etal-2020-sigmorphon}, which contains 90.
Figure~\ref{fig:map} shows where these languages are primarily spoken according to \ac{wals} \cite{dryer2013wals}. 
We observe a skew towards languages spoken in Europe, where multiple languages are used in more than 50 papers, while not a single language from the Americas is included in more than 10 papers.

\subsection{Justifications}
\label{sec:just}
The papers include a wide spectrum of justifications for their claims.
A large portion provide no justification at all.
Some use genealogy: the selection of 24 languages from \citet{xu-etal-2022-cross} aims to cover ``a reasonable variety of language families'', while the dataset created by \citet{zhang-etal-2023-miracl} consists of ``[18] languages (\ldots) from 10 language families and 13 sub-families''.
Others use a selection of typological features, for instance, \citet{mott-etal-2020-morphological} mention that ``the nine languages in our corpus cover five primary language families (\ldots), and a range of morphological phenomena (\ldots)''.
Some also mention typological databases in their language selection: \citet{muradoglu-hulden-2022-eeny} consider ``languages that exhibit varying degrees of complexity for inflection. We also consider morphological characteristics coded in \ac{wals} (\ldots)''.
A rather systematic approach to language selection is found in \citet{jancso-etal-2020-acqdiv}.
They use a clustering algorithm on vectors with features from two typological databases to find the most distant clusters to sample languages from.

Still, most other papers justify their typological diversity in a `post-hoc' way.
They generally do not mention \textit{initial} language selection considerations.
Rather, they  mention how diverse the sample is \emph{after} it has been created. 
The XTREME-R dataset \cite{ruder-etal-2021-xtreme}, which we will discuss further in \S \ref{sec:model_eval}, is exemplary of this.
They mention `diversity indices' that cover family, as well as typology and compare theirs to other datasets, but do not detail how or if these were used in sampling. 

\section{Typological Analysis}
\label{sec:analysis}

\begin{figure*}[t]
    \centering
    \includegraphics[width=0.45\textwidth]{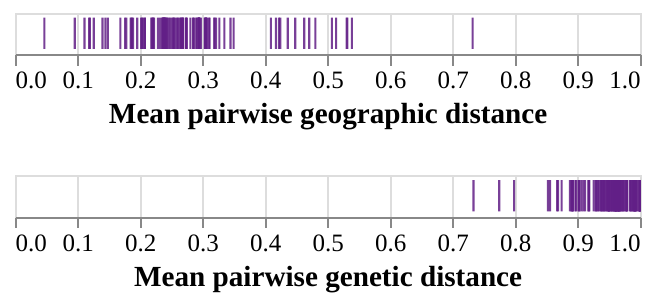}
    \hspace{3em}
    \includegraphics[width=0.45\textwidth]{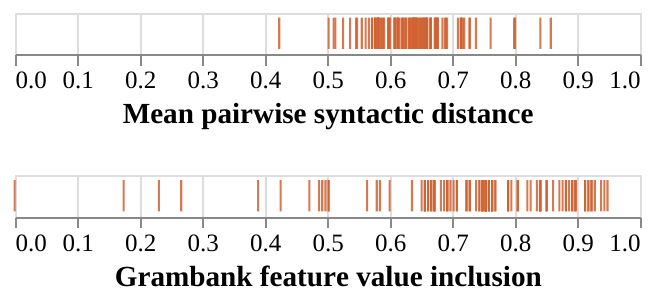}
    \caption{Distributions of mean pairwise \texttt{lang2vec} distances and feature inclusion per paper. On the \textcolor{low-color}{left} are approximations based on common justifications for claiming `typological diversity': geography ($\mu=0.28$, $\sigma=0.11$) and genealogy ($\mu=0.94$, $\sigma=0.05$). On the \textcolor{high-color}{right} two different approximations based on typological features: MPSD ($\mu=0.64$, $\sigma=0.07$) and Grambank feature value inclusion ($\mu=0.72$, $\sigma=0.17$).}
    \label{fig:lang2vec_mpd}
\end{figure*}

From Section \ref{sec:results-languages}, it follows that there is a geographical skew, where languages from certain areas are over-represented (Figure~\ref{fig:map}).
This is not surprising, as it is often more feasible to select languages for which there are existing resources and typological descriptions, than languages for which these have to be gathered from scratch. 
This constitutes bibliographic bias \citep{rijkhoff1998language}.
However, we cannot draw conclusions about typological diversity on this basis, as typology is not equal to geography, nor phylogeny \citep{cysouw2013disentangling}.
Therefore, in this section, we analyze the typological diversity according to metrics that take into account syntactic features, and the absolute included typological feature values.
\citet{ponti-etal-2020-xcopa} previously quantified language diversity with \textit{typology} (URIEL), \textit{geography}, and \textit{family} indices.
These metrics do not take into account the distance between pairs of languages, and thus do not provide insights into the independence of languages in a sample.
Additionally, no code or clear descriptions are provided, which limits their use.

\newpage

Our metrics\footnote{Formal definitions are provided in Appendix \ref{app:metrics}.} include  pairwise language distances (\S \ref{sec:dist}) and absolute typological feature value inclusion (\S \ref{sec:cov}) across papers.
Lastly, we look into the relationship between the number of included languages and typological diversity, to help guide future language selection efforts (\S \ref{sec:num_langs}).

\subsection{Mean Language Distance}
\label{sec:dist}
We first approximate the typological diversity of each paper's language selection with \texttt{lang2vec} \citep{littell-etal-2017-uriel}.
This toolkit provides language distances that are calculated based on language vectors from URIEL, a resource that contains information from a range of typological databases, including \ac{wals}.
We choose \texttt{lang2vec} specifically, because its aggregated information enables analysis across databases, and because its vectorized format  mitigates issues stemming from incomplete feature coverage in typological databases.

For each paper in our survey that claims a typologically diverse language set and that is covered in \texttt{lang2vec} (312 / 315 unique languages), we calculate the \textsc{mean pairwise syntactic distance (MPSD)} for the included languages.
That is, we take the average of all pre-computed syntactic \texttt{lang2vec} distances for each pair of languages in a given paper's selection. 
We only take into account languages that have at least 5\% coverage in the URIEL vectors.
We do this to make sure the very low coverage languages do not distort the results.
This affects 31 papers in our dataset, resulting in a total of 242 unique languages.
The top-right plot in Figure~\ref{fig:lang2vec_mpd} shows the distribution of MPSDs across papers.\footnote{To illustrate this: the distance between Danish and Norwegian (same family, same genus) is 0.22, Danish-Spanish (same family, different genus) is 0.59 and Danish-Japanese (different family, different genus) is 0.69.} %
We compare our typology-based approximations with mean geographical and genetic pairwise distances.
Interestingly, the genetic distance is much higher and less spread out than the syntactic distance.
This emphasizes that genealogical sampling does not by default ensure typological diversity.

The typological distances vary considerably, with outliers on either side (Figure~\ref{fig:lang2vec_mpd}).
In papers claiming typological diversity, the minimum MPSD is found in \citet{goel-etal-2022-unsupervised}, who use English, French, and Spanish.
The maximum MPSD is found for \citet{vania-etal-2019-systematic} their selection of North Sámi, Galician, and Kazah.

\subsection{Inclusion of Typological Features}
\label{sec:cov}
Because \texttt{lang2vec} distance calculations are based on feature vectors, our previous analysis does not provide information regarding \textit{which} typological feature values are actually included in the language set.
For example, it does not tell us which word order variations are covered in a given language selection.
This information can be useful though, because covering more typological feature values in an evaluation set means that robustness towards previously unseen typological characteristics is increased.
Following the \textit{saturation} measure by \citet{miestamo2016},
we look into the inclusion of individual typological feature values per language selection from each paper. We use the Grambank\footnote{\url{https://grambank.clld.org/}} \citep{skirgaard2023dataset} database, because it has high coverage for grammatical features and is currently actively maintained.
For each feature in Grambank, we count whether all possible, non-missing feature values are represented in the paper's language selection.\footnote{Note that for our purposes, we treat ? as a missing value.}
That is, for a feature such as \textit{GB020: Are there definite or specific articles?}, we count whether all non-missing options are represented by the particular set of languages from a paper and divide this by the total number of features. 
The spread of these values is shown in the bottom right graph in Figure~\ref{fig:lang2vec_mpd}.
We treat missing languages the same as in the MPSD calculations; we discard them during calculation, but we do not discard the entire paper.
Twelve papers are affected by this.
The average Grambank feature value inclusion is 0.73.

\begin{figure}[t]
    \centering
    \includegraphics[width=\columnwidth]{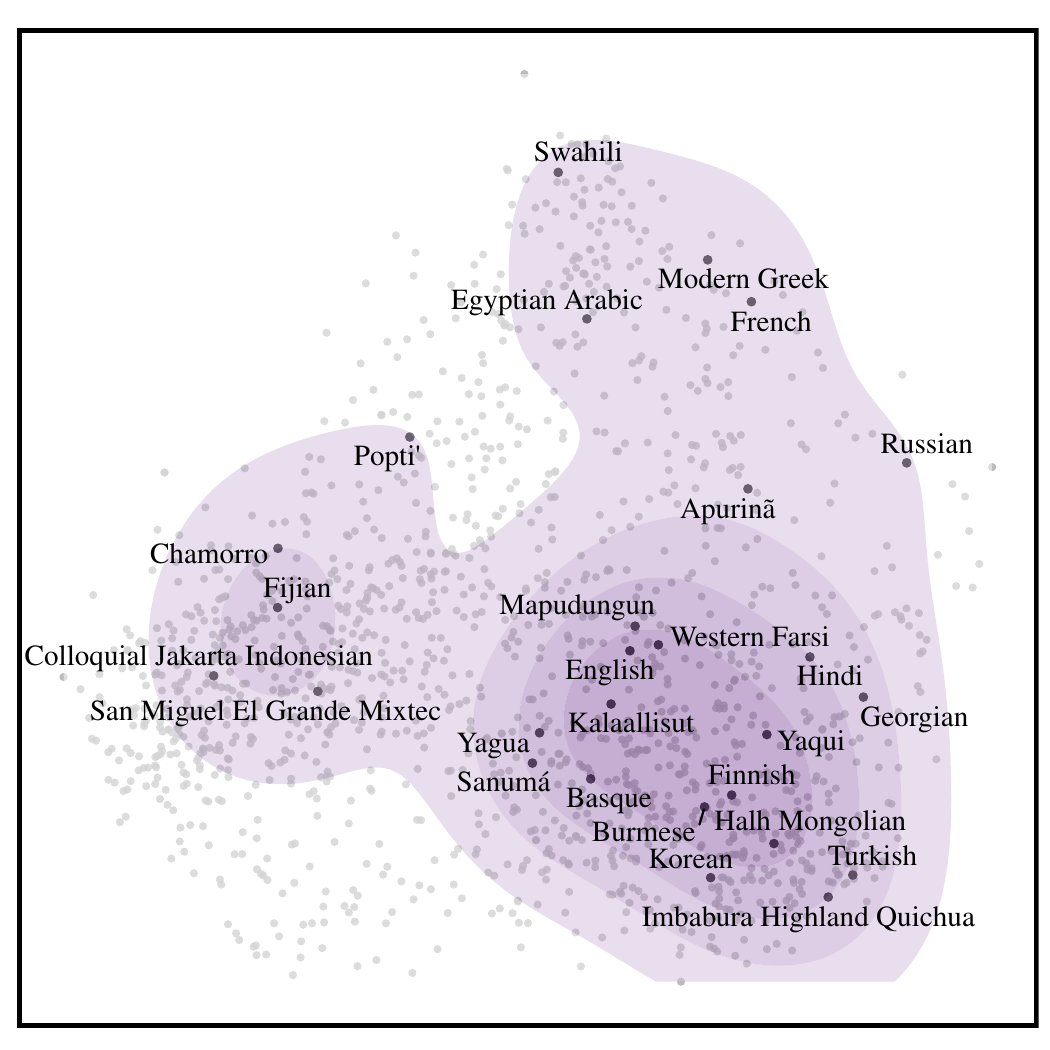}
    \caption{PCA plot for the paper with the highest absolute feature coverage \citep{gutierrez-vasques-etal-2021-characters}, where each point represents a language in Grambank.}
    \label{fig:pca}
\end{figure}

While the accumulation of absolute feature values covered in a language selection provides useful robustness insights, it does not capture that languages consist of \textit{combinations} of features that are not entirely independent.
\citet{skirgaard2023grambank} show how language families can reliably be visualized as distinct groupings in Grambank's typological design space using principle component analysis (PCA).
Similarly, in Figure~\ref{fig:pca} we show the PCA plot for the paper that has the highest Grambank feature inclusion of all papers: \citet{gutierrez-vasques-etal-2021-characters}.
While their language sample covers much of the Grambank's typological space, we observe a skew towards certain languages (bottom right), and a lack of representation for others (e.g., bottom left).
This means that even the paper with the highest feature value inclusion in our survey contains challenges when it comes to complete and fair evaluation.

\subsection{Number of Languages}
\label{sec:num_langs}
As shown in Figure~\ref{fig:num_langs}, the number of included languages varies across papers.
Should one aim to evaluate on as many languages as possible? What is the effect of selecting many (similar) languages?
Figure~\ref{fig:typdiv_numlangs} and \ref{fig:typcov_numlangs} show the relationship between the number of languages and the MPSDs, and the absolute typological feature value inclusion, respectively.
We find that smaller language sets (0-10 languages) exhibit considerable variation in terms of average syntactic distances, while the larger language sets (30+ languages) seem consistent between 0.6 and 0.7. 
This implies that adding more languages does not necessarily raise the average syntactic distance much.
In \ac{nlp}, adding more languages typically means adding more \emph{similar} languages, since it is easier to incorporate existing datasets than it is to create new ones.
As a result, the average performance will also be skewed towards these similar languages in evaluation (see \S \ref{sec:model_eval}).
Furthermore, we observe that up to a certain point, including more languages implies that more typological feature values are covered. However, this increase flattens at approximately 40 languages.
Importantly, this highlights the fact that simply adding more languages to an experimental study is not by itself a contribution in terms of typological generalizability -- one must take care \emph{which} languages are included.
To illustrate this, we highlight the MKQA dataset \cite{longpre-etal-2021-mkqa}.
This dataset includes 25 languages, more than twice the median, but has a rather average MPSD of 0.61 (and a Grambank feature inclusion of 0.89).
This is in part due to its inclusion of several typologically similar languages: \{English, German, Dutch\}, \{Cantonese, Mandarin\}, \{Spanish, Portuguese\} and \{Swedish, Danish, Norwegian\}.

\begin{figure}[t]
     \centering
     \begin{subfigure}[b]{0.49\columnwidth}
        \centering
        \includegraphics[width=\columnwidth]{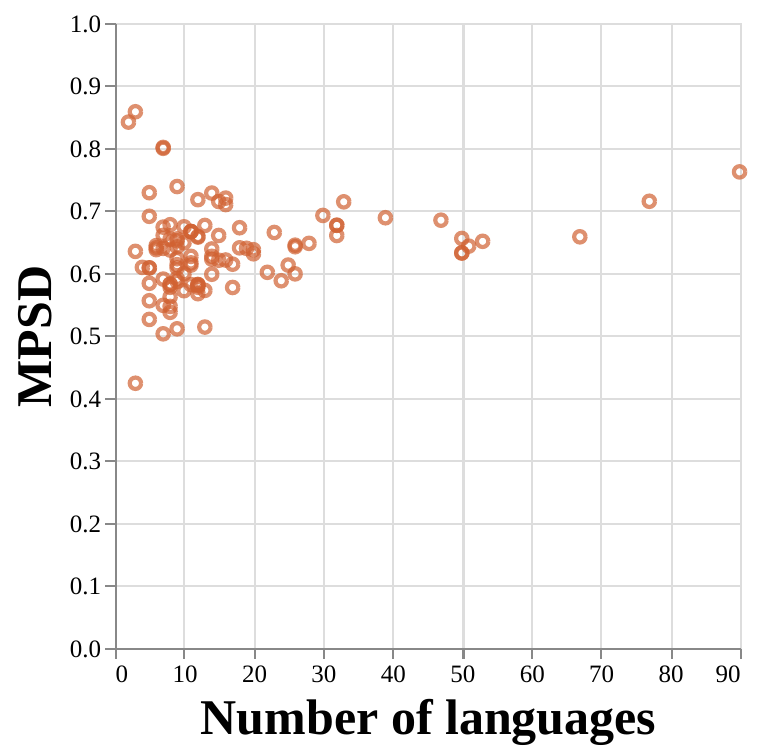}
        \phantomsubcaption
        \label{fig:typdiv_numlangs}
     \end{subfigure}
     \hfill
     \begin{subfigure}[b]{0.49\columnwidth}
            \centering
            \includegraphics[width=\columnwidth]{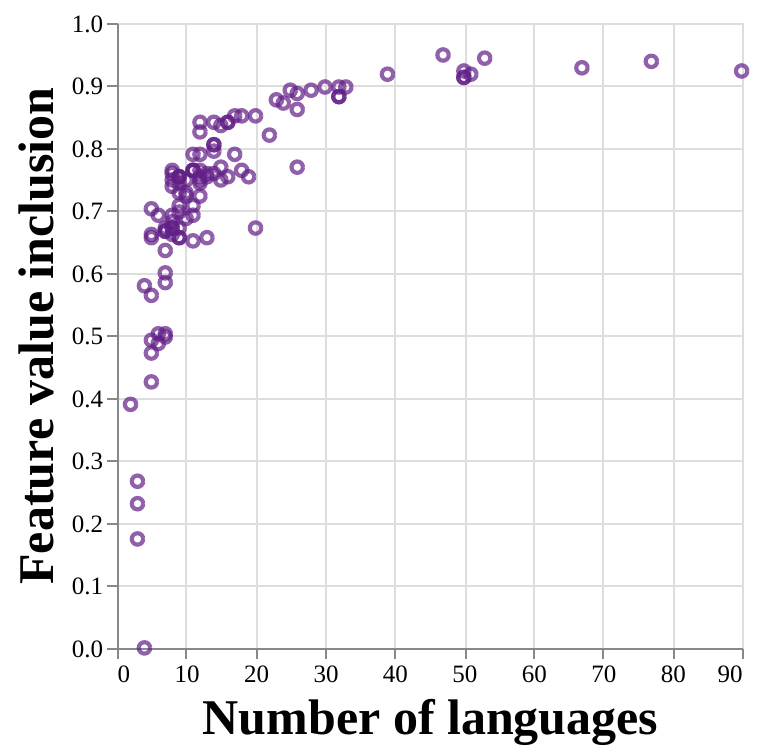}
            \phantomsubcaption
            \label{fig:typcov_numlangs}
     \end{subfigure}
    \caption{MPSD (a) and Grambank feature value inclusion (b) per paper by number of languages.}
    \label{fig:cov}
\end{figure}

\begin{table*}[ht]
    \centering
    \tiny
    \begin{tabular}{ll||llr|lllllll}
	\toprule
	Subtask                                 & Model   & \textbf{Overall}    & \textbf{By F}       & $\Delta$       & \makecell{Strong\\ Pre} & \makecell{Weak\\ Pre}                          & \makecell{Equal\\ Pre \& Suf}                  & \makecell{Strong\\ Suf}                          & \makecell{Weak\\ Suf}                          & \makecell{Little\\ Aff}                        & NA                                           \\\midrule\midrule
	\multirow{2}{*}{\textbf{Mewsli-X}\map}  & XLM-R-L & 45.75 (\textit{11}) & 36.23 (\textit{11}) & \tgrad{-9.52}  & - (\textit{0})        & - (\textit{0})                               & - (\textit{0})                               & \cellcolor{high-color!40}  47.86 (\textit{10}) & \cellcolor{low-color!40}  24.60 (\textit{1}) & - (\textit{0})                               & - (\textit{0})                               \\
	                                        & mBERT   & 38.58 (\textit{11}) & 27.29 (\textit{11}) & \tgrad{-11.29} & - (\textit{0})        & - (\textit{0})                               & - (\textit{0})                               & \cellcolor{high-color!40}  41.09 (\textit{10}) & \cellcolor{low-color!40}  13.50 (\textit{1}) & - (\textit{0})                               & - (\textit{0})                               \\\midrule
	\multirow{3}{*}{\textbf{XNLI}\acc}      & XLM-R   & 79.24 (\textit{15}) & 76.54 (\textit{15}) & \tgrad{-2.70}  & - (\textit{0})        & \cellcolor{low-color!40}  71.20 (\textit{1}) & - (\textit{0})                               & \cellcolor{high-color!40}  80.06 (\textit{12}) & - (\textit{0})                               & 78.35 (\textit{2})                           & - (\textit{0})                               \\
	                                        & mBERT   & 66.51 (\textit{15}) & 60.17 (\textit{15}) & \tgrad{-6.35}  & - (\textit{0})        & \cellcolor{low-color!40}  49.30 (\textit{1}) & - (\textit{0})                               & \cellcolor{high-color!40}  68.60 (\textit{12}) & - (\textit{0})                               & 62.60 (\textit{2})                           & - (\textit{0})                               \\
	                                        & mT5     & 84.85 (\textit{15}) & 82.92 (\textit{15}) & \tgrad{-1.92}  & - (\textit{0})        & \cellcolor{low-color!40}  80.60 (\textit{1}) & - (\textit{0})                               & \cellcolor{high-color!40}  85.57 (\textit{12}) & - (\textit{0})                               & 82.60 (\textit{2})                           & - (\textit{0})                               \\\midrule
	\multirow{2}{*}{\textbf{LAReQA}\map}    & XLM-R-L & 40.75 (\textit{11}) & 40.54 (\textit{11}) & \tgrad{-0.22}  & - (\textit{0})        & - (\textit{0})                               & - (\textit{0})                               & \cellcolor{high-color!40}  40.88 (\textit{9})  & - (\textit{0})                               & \cellcolor{low-color!40}  40.20 (\textit{2}) & - (\textit{0})                               \\
	                                        & mBERT   & 21.58 (\textit{11}) & 19.24 (\textit{11}) & \tgrad{-2.35}  & - (\textit{0})        & - (\textit{0})                               & - (\textit{0})                               & \cellcolor{high-color!40}  22.92 (\textit{9})  & - (\textit{0})                               & \cellcolor{low-color!40}  15.55 (\textit{2}) & - (\textit{0})                               \\\midrule
	\multirow{3}{*}{\textbf{XQuAD}\f}       & XLM-R-L & 77.21 (\textit{11}) & 77.24 (\textit{11}) & \tgrad{+0.04}  & - (\textit{0})        & - (\textit{0})                               & - (\textit{0})                               & \cellcolor{high-color!40}  77.19 (\textit{9})  & - (\textit{0})                               & \cellcolor{low-color!40}  77.30 (\textit{2}) & - (\textit{0})                               \\
	                                        & mBERT   & 65.05 (\textit{11}) & 61.84 (\textit{11}) & \tgrad{-3.21}  & - (\textit{0})        & - (\textit{0})                               & - (\textit{0})                               & \cellcolor{high-color!40}  66.89 (\textit{9})  & - (\textit{0})                               & \cellcolor{low-color!40}  56.80 (\textit{2}) & - (\textit{0})                               \\
	                                        & mT5     & 81.54 (\textit{11}) & 80.55 (\textit{11}) & \tgrad{-0.99}  & - (\textit{0})        & - (\textit{0})                               & - (\textit{0})                               & \cellcolor{high-color!40}  82.10 (\textit{9})  & - (\textit{0})                               & \cellcolor{low-color!40}  79.00 (\textit{2}) & - (\textit{0})                               \\\midrule
	\multirow{3}{*}{\textbf{MLQA}\f}        & XLM-R-L & 72.71 (\textit{7})  & 73.33 (\textit{7})  & \tgrad{+0.62}  & - (\textit{0})        & - (\textit{0})                               & - (\textit{0})                               & \cellcolor{high-color!40}  72.47 (\textit{6})  & - (\textit{0})                               & \cellcolor{low-color!40}  74.20 (\textit{1}) & - (\textit{0})                               \\
	                                        & mBERT   & 61.30 (\textit{7})  & 60.84 (\textit{7})  & \tgrad{-0.46}  & - (\textit{0})        & - (\textit{0})                               & - (\textit{0})                               & \cellcolor{high-color!40}  61.48 (\textit{6})  & - (\textit{0})                               & \cellcolor{low-color!40}  60.20 (\textit{1}) & - (\textit{0})                               \\
	                                        & mT5     & 75.59 (\textit{7})  & 75.97 (\textit{7})  & \tgrad{+0.38}  & - (\textit{0})        & - (\textit{0})                               & - (\textit{0})                               & \cellcolor{high-color!40}  75.43 (\textit{6})  & - (\textit{0})                               & \cellcolor{low-color!40}  76.50 (\textit{1}) & - (\textit{0})                               \\
	\midrule\midrule
	\multirow{2}{*}{\textbf{Tatoeba}\acc}   & XLM-R   & 77.29 (\textit{41}) & 64.92 (\textit{36}) & \tgrad{-12.36} & - (\textit{0})        & \cellcolor{low-color!40}  31.30 (\textit{1}) & \cellcolor{low-color!40}  58.60 (\textit{1}) & \cellcolor{high-color!40}  82.10 (\textit{28}) & 76.37 (\textit{3})                           & 77.43 (\textit{3})                           & 63.74 (\textit{5})                           \\
	                                        & mBERT   & 43.33 (\textit{41}) & 32.03 (\textit{36}) & \tgrad{-11.30} & - (\textit{0})        & \cellcolor{low-color!40}  12.10 (\textit{1}) & \cellcolor{low-color!40}  31.00 (\textit{1}) & \cellcolor{high-color!40}  49.24 (\textit{28}) & 39.27 (\textit{3})                           & 32.90 (\textit{3})                           & 27.68 (\textit{5})                           \\\midrule
	\multirow{2}{*}{\textbf{UD-POS}\f}      & XLM-R-L & 74.96 (\textit{38}) & 71.12 (\textit{36}) & \tgrad{-3.84}  & - (\textit{0})        & - (\textit{0})                               & \cellcolor{low-color!40}  74.30 (\textit{1}) & \cellcolor{high-color!40}  79.75 (\textit{28}) & 71.05 (\textit{2})                           & 45.98 (\textit{5})                           & 84.50 (\textit{2})                           \\
	                                        & mBERT   & 70.90 (\textit{38}) & 64.43 (\textit{36}) & \tgrad{-6.47}  & - (\textit{0})        & - (\textit{0})                               & \cellcolor{low-color!40}  59.30 (\textit{1}) & \cellcolor{high-color!40}  75.51 (\textit{28}) & 60.75 (\textit{2})                           & 48.66 (\textit{5})                           & 77.95 (\textit{2})                           \\\midrule
	\multirow{3}{*}{\textbf{XCOPA}\acc}     & XLM-R   & 69.22 (\textit{11}) & 65.93 (\textit{9})  & \tgrad{-3.28}  & - (\textit{0})        & \cellcolor{low-color!40}  61.80 (\textit{1}) & - (\textit{0})                               & \cellcolor{high-color!40}  73.93 (\textit{6})  & - (\textit{0})                               & 75.30 (\textit{2})                           & 52.70 (\textit{2})                           \\
	                                        & mBERT   & 56.05 (\textit{11}) & 54.75 (\textit{9})  & \tgrad{-1.30}  & - (\textit{0})        & \cellcolor{low-color!40}  52.20 (\textit{1}) & - (\textit{0})                               & \cellcolor{high-color!40}  57.70 (\textit{6})  & - (\textit{0})                               & 56.20 (\textit{2})                           & 52.90 (\textit{2})                           \\
	                                        & mT5     & 74.89 (\textit{11}) & 73.24 (\textit{9})  & \tgrad{-1.65}  & - (\textit{0})        & \cellcolor{low-color!40}  74.10 (\textit{1}) & - (\textit{0})                               & \cellcolor{high-color!40}  78.00 (\textit{6})  & - (\textit{0})                               & 77.60 (\textit{2})                           & 63.25 (\textit{2})                           \\\midrule
	\multirow{2}{*}{\textbf{WikiANN-NER}\f} & XLM-R-L & 64.43 (\textit{48}) & 62.02 (\textit{40}) & \tgrad{-2.41}  & - (\textit{0})        & \cellcolor{low-color!40}  69.90 (\textit{1}) & \cellcolor{low-color!40}  62.10 (\textit{1}) & \cellcolor{high-color!40}  66.92 (\textit{31}) & 61.37 (\textit{3})                           & 48.17 (\textit{4})                           & 63.66 (\textit{8})                           \\
	                                        & mBERT   & 62.68 (\textit{48}) & 61.73 (\textit{40}) & \tgrad{-0.95}  & - (\textit{0})        & \cellcolor{low-color!40}  72.70 (\textit{1}) & \cellcolor{low-color!40}  65.00 (\textit{1}) & \cellcolor{high-color!40}  64.93 (\textit{31}) & 57.23 (\textit{3})                           & 49.38 (\textit{4})                           & 61.12 (\textit{8})                           \\\midrule
	\multirow{3}{*}{\textbf{TyDiQA}\f}      & XLM-R-L & 64.29 (\textit{9})  & 62.57 (\textit{8})  & \tgrad{-1.72}  & - (\textit{0})        & \cellcolor{low-color!40}  66.40 (\textit{1}) & - (\textit{0})                               & \cellcolor{high-color!40}  65.67 (\textit{6})  & - (\textit{0})                               & \cellcolor{low-color!40}  59.10 (\textit{1}) & \cellcolor{low-color!40}  59.10 (\textit{1}) \\
	                                        & mBERT   & 58.36 (\textit{9})  & 55.09 (\textit{8})  & \tgrad{-3.26}  & - (\textit{0})        & \cellcolor{low-color!40}  59.70 (\textit{1}) & - (\textit{0})                               & \cellcolor{high-color!40}  60.97 (\textit{6})  & - (\textit{0})                               & \cellcolor{low-color!40}  46.20 (\textit{1}) & \cellcolor{low-color!40}  53.50 (\textit{1}) \\
	                                        & mT5     & 81.94 (\textit{9})  & 83.73 (\textit{8})  & \tgrad{+1.78}  & - (\textit{0})        & \cellcolor{low-color!40}  87.20 (\textit{1}) & - (\textit{0})                               & \cellcolor{high-color!40}  80.52 (\textit{6})  & - (\textit{0})                               & \cellcolor{low-color!40}  83.60 (\textit{1}) & \cellcolor{low-color!40}  83.60 (\textit{1}) \\
	\bottomrule
\end{tabular}

    \caption{XTREME-R results grouped by inflection type (WALS 26A). 
    \textbf{Overall} refers to the average over all languages. 
    \textbf{By~F}eature is the average of the WALS feature averages, excluding languages for which there is no coverage. 
    The delta shows the difference of \textbf{Overall} and \textbf{By F}. 
    Tasks that have coverage for all their included languages are on top, those partially covered are in the bottom portion. Morphological inflection types are: \emph{Aff} = affixing, \emph{Suf} = suffixing, \emph{Pre} = prefixing and \emph{NA} referring to \emph{Not Available} (in WALS). The \textcolor{high-color}{highest} and the \textcolor{low-color}{lowest} non-zero number of languages per grouping are highlighted. The \emph{(italicized)} number refers to the number of languages in a particular subset. Metrics: \acctext~=~Accuracy, \maptext~=~mAP@20, \ftext~=~F1.}
    \label{tab:inflectional}
\end{table*}
\section{Downstream Implications}
\label{sec:model_eval}

Inconsistencies in `typologically diverse' language selection can have an effect on downstream evaluation.
While reporting all metrics for all languages separately is preferable, space constraints often lead to reporting \textit{averages} across languages in multilingual evaluation.
However, \citet{blog} demonstrates that when a language set contains a skew towards a certain language family, simply taking the micro average over all languages for evaluation gives an overestimation of performance.
While family can serve as a proxy for typological diversity, we here present an analysis based directly on typological characteristics. %
In this way, we gain more fine-grained insights into the effects of typologically skewed language selection for model evaluation. 

Specifically, we use \mbox{XTREME-R} \cite{ruder-etal-2021-xtreme}, a popular\footnote{The original XTREME dataset has almost 800 citations as of writing, XTREME-R, which is an extension of the original, more than 100.} `typologically diverse' multilingual benchmark covering several tasks and domains.
\mbox{XTREME-R} contains and expands upon ten existing datasets.
As a result, not all tasks are available in all languages; while there are 50 unique languages in total, the number of languages per task ranges from 7 to 48.\footnote{See Appendix \ref{app:pca}, Figure~\ref{fig:xtremer-pca}.}
We group these languages by the `Prefixing vs. Suffixing in Inflectional Morphology' (26A) and `Order of Subject, Object and Verb' (81A) features, as provided by \ac{wals} \citep{wals-26,wals-81}.
These features have high language coverage, which makes our analysis as comprehensive as possible. 
Table~\ref{tab:inflectional} shows groupings and coverage of the inflectional feature (26A).
The word order analysis is included in Appendix \ref{sec:app-word-order}, Table~\ref{tab:word-order}.

Firstly, we see that XTREME-R contains a skew in terms of morphological inflection. 
For all tasks, the majority of languages are \textit{strongly suffixing} (see orange cells).
Some feature values, such as \textit{equal prefixing and suffixing} and \textit{weak prefixing} are underrepresented, with at most one language per subtask.
Remarkably, \textit{strong prefixing} does not appear in XTREME-R at all.
This implies that one should be careful with the \textit{implication} of generalizability that evaluating on a `typologically diverse' dataset gives.

The delta column shows the difference between the macro average over all languages and the macro average per feature value.
Here, we observe similar patterns as in \citet{blog}, namely that estimations of multilingual model performance vary considerably when accounting for typological imbalances. %
We suggest that drawing and implying generalizable conclusions about typologically diverse evaluation should ideally be supported by showing that performance holds across a range of typological properties.

\section{Insights and Recommendations}
We systematically analyze claims of `typological diversity' in \ac{nlp} research.
We approximate diversity in terms of average syntactic language distance (MPSD) and absolute typological feature value inclusion in Grambank.
Our analysis shows that (1) there is no consistent definition or methodology when making `typological diversity' claims, (2) estimates of typological diversity exhibit a considerable variation across papers and (3) aggregated results can give distorted views of multilingual performance estimates.

We recommend future approaches to include an operationalization of `typological diversity' when making such claims.
This can be in terms of generalizability or regarding a particular typological phenomenon of interest.
Additionally, we recommend adding an empirical justification, especially when claims relate to generalizability.
The feature value inclusion, MPSD or PCA plots we have shown are examples of these justifications.
Including these has the potential to benefit multilingual NLP, as it enables more fine-grained insights into typological challenges of language modelling.

\section*{Limitations}
Our measures of typological diversity are approximations.
There is no typological database that covers every aspect of every language.
This also ties into the bibliographic bias of typological databases in general.
In addition, languages are not necessarily as discrete as they appear in typological resources either \cite{levshina2023why,baylor-etal-2024-multilingual}.
Nonetheless, we believe that the reporting of typological diversity can be more principled than it currently is, despite incomplete resources.

Another limitation is that our search is based only on the title and abstract of papers.
There are almost certainly papers that contain a claim in other sections.
However, we chose this route since we wanted to (1) cast a wide net, (2) perform some pre-filtering of papers, assuming that a mention of `typological diversity' in the title or abstract is a good indicator that it is a prominent aspect of that paper, and (3) make sure that the data we could get was of good quality, which is a lot harder with (old) pdf files for example.

Our arguments for assessing linguistic typology directly in NLP, rather than phylogeny and geography as proxies, relates to claims of `typological diversity'. We acknowledge that geographical and genealogical information could be useful for other purposes. %

\section*{Ethics Statement}
In this paper, we investigate the use of the term `typological diversity' in multilingual NLP research.
We do not make a claim that NLP applications \textit{should} be expanded to as many languages as possible, as introducing these technologies are not necessarily always a positive influence \citep{bird-2020-decolonising}.
Instead, our aim is to gain a common understanding and clear explanations regarding the usage of these claims.

\section*{Acknowledgements}
EP and JB are funded by the Carlsberg Foundation, under the Semper Ardens:~Accelerate programme (project nr.~CF21-0454). WP is funded by a KU Leuven Bijzonder Onderzoeksfonds C1 project with reference C14/23/096.

We thank the TypNLP group at Aalborg University and the LAGoM-NLP group at KU Leuven for insightful feedback on earlier versions of this paper, in particular Heather Lent.
We also thank Thomas Bauwens for his help in creating the tables (using \texttt{fiject}).

\bibliography{anthology,custom}

\appendix

\newpage
\onecolumn

\section{Paper Retrieval Details}\label{sec:app-paper-details}

\begin{table}[h]
    \centering
    \small
    \begin{tabular}{lrrrrrr}
        \toprule
        \textbf{Venue} ($\downarrow$)        & \textbf{Date range} & \textbf{\# papers}   & \textbf{\# search hits} & \textbf{\# \texttt{has\_claim}} & \textbf{\# \texttt{introduces\_dataset}} & \textbf{Date retrieved }\\
        \midrule
        AAAI          & 1980--2023 & \num{18370} & \num{2}        & \num{2}                & \num{0}                         & 2024-01-02     \\
        ACL Anthology & 1952--2023 & \num{92006} & \num{188}      & \num{105}              & \num{36}                        & 2023-12-11     \\
        ICLR          & 2013--2023 & \num{5307}  & \num{3}        & \num{2}                & \num{1}                         & 2024-01-02     \\
        ICML          & 2011--2023 & \num{8372}  & \num{1}        & \num{1}                & \num{1}                         & 2024-01-02     \\
        IJCAI         & 2007--2023 & \num{8911}  & \num{0}        & \num{0}                & \num{0}                         & 2024-01-03     \\
        NeurIPS       & 1987--2023 & \num{20343} & \num{0}        & \num{0}                & \num{0}                         & 2023-12-23     \\
        \midrule
        \textbf{Total} & 1952-2023 & \num{153309} & \num{194} & \num{110} & \num{38} & - \\
        \bottomrule
    \end{tabular}
    \caption{Detailed description of the dataset collected for annotation. Due to availability and quality reasons, not all proceedings of all years for all venues could be retrieved.}
    \label{tab:dataset}
\end{table}

\section{Diversity Metrics}\label{app:metrics}
\paragraph{Mean Pairwise Distance (MPD)} For a language sample $L$ and feature vectors $V$, we retrieve $v_l$ for $l \in L$.
We then calculate the mean, pairwise Euclidean distances between all pairs in $L$:
\begin{displaymath}
    MPD(L) = \frac{1}{|L|(|L|-1)} \sum_{i, j \in L, i \neq j} \lVert v_i - v_j \rVert_2
\end{displaymath}

where the feature vectors from $V$ consists of the \textsc{geography}, \textsc{genetic} or \textsc{syntactic} vectors in \texttt{lang2vec} (as shown per paper in Figure~\ref{fig:lang2vec_mpd}).

\paragraph{Feature Value Inclusion (FVI)} For a language sample $L$ and feature vectors $V$ we look at every $v_l$ for $l \in L$. We check if all possible, non-missing values (i.e., $x \not\in \{\text{no\_cov}, ?\}$, the values indicating `no coverage' and `unsure' in Grambank) for a particular feature are covered ($\mathcal{V}_{F_i}$). We do this for the set of all features in Grambank $\mathcal{F}$. Finally, we take the average over the total number of features:
\begin{displaymath}
    FVI(L) = \frac{1}{|\mathcal{F}|} \sum_{i=1}^{|\mathcal{F}|} \frac{|\{v_l[F_i] : l \in L\}|}{|\mathcal{V}_{F_i}|}
\end{displaymath}

\clearpage

\section{Long Tail of Included Languages}
\label{sec:app-long-tail}
\begin{figure}[h]
    \captionsetup[sub]{belowskip=0.2em}
    \begin{subfigure}[b]{\textwidth}
        \centering
        \includegraphics[width=\textwidth]{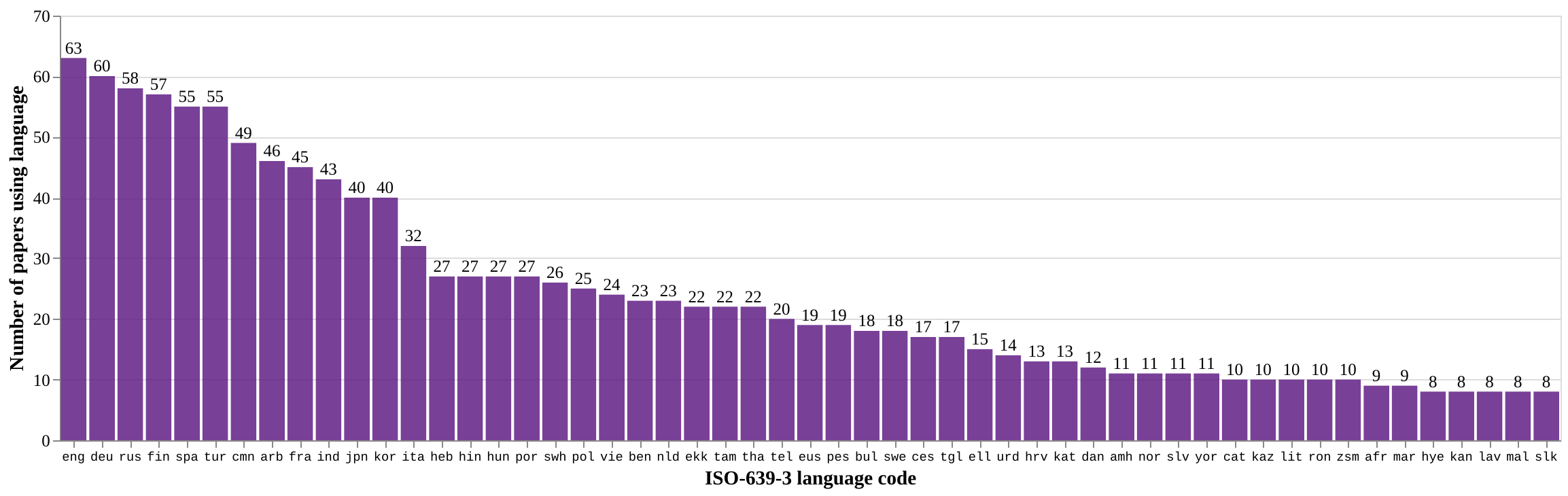}
        \caption{Top portion of the long tail.}
        \label{fig:long_tail_top}
     \end{subfigure}
     
    \begin{subfigure}[b]{\textwidth}
        \centering
        \includegraphics[width=\textwidth]{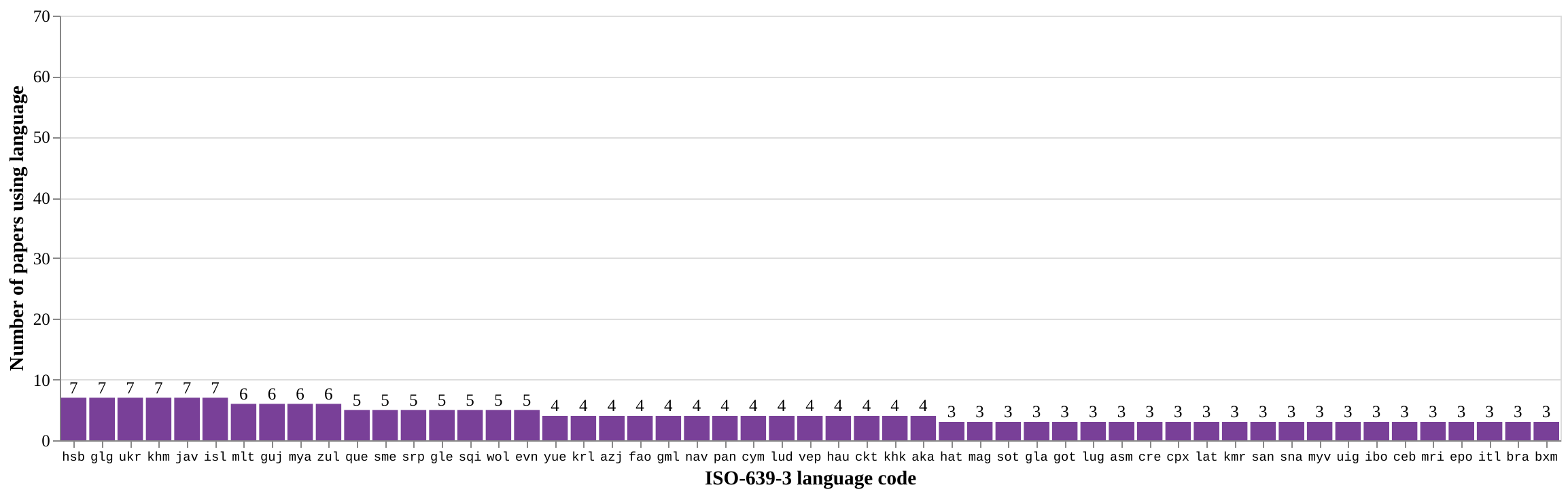}
        \caption{Continued tail of included languages (not exhaustive).}
        \label{fig:long_tail_bottom}
    \end{subfigure}
    \caption{Long tail of included languages.}
    \label{fig:long_tail}
\end{figure}

\clearpage

\section{`Typologically Diverse' Datasets}
\label{app:datasets}

\begin{table}[h]
    \centering
    \footnotesize
    \begin{tabular}{llrlll}
	\toprule
	\textbf{Paper}                                & \textbf{Dataset name}  & \textbf{$|$L$|$ ($\downarrow$)} & \textbf{MPSD}                  & \textbf{FVI}                   & \textbf{Task(s) or Topic}           \\\midrule
	\citet{vylomova-etal-2020-sigmorphon}         &                        & 90                              & 0.76\lowcov                   & 0.92\missing                  & Morphological inflection            \\
	\citet{henrichsen-uneson-2012-smallworlds}    & SMALLWorlds            & 53                              & 0.65\lowcov                   & \cellcolor{high-color!40}0.94 & ASR                                 \\
	\citet{fitzgerald-etal-2023-massive}          & MASSIVE                & 51                              & 0.64\lowcov                   & 0.92                          & Slot-filling, Intent classification \\
	\citet{ruder-etal-2021-xtreme}                & XTREME-R               & 50                              & 0.66\lowcov                   & 0.92\missing                  & Classification, Parsing, IR, QA     \\
	\citet{kodner-etal-2022-sigmorphon}           &                        & 32\nolang                       & 0.68\lowcov                   & 0.88\missing                  & Morphological inflection            \\
	\citet{goldman-etal-2023-sigmorphon}          &                        & 26                              & 0.64                          & 0.89                          & Morphological inflection            \\
	\citet{longpre-etal-2021-mkqa}                & MKQA                   & 25                              & 0.61                          & 0.89                          & QA                                  \\
	\citet{jiang-etal-2020-x}                     & X-FACTR                & 23                              & 0.66\lowcov                   & 0.88                          & Knowledge retrieval                 \\
	\citet{bugliarello2022iglue}                  & IGLUE                  & 20                              & 0.63                          & 0.85                          & Visual QA, IR, RE, NLI              \\
	\citet{dione-etal-2023-masakhapos}            & MasakhaPOS             & 20                              & 0.64\lowcov                   & 0.67                          & Part-of-speech tagging              \\
	\citet{zhang-etal-2023-miracl}                & MIRACL                 & 18                              & 0.64                          & 0.85                          & IR                                  \\
	\citet{palma-gomez-etal-2023-using}           &                        & 16                              & 0.62\lowcov                   & 0.84                          & Language learning, cloze test       \\
	\citet{asai-etal-2022-mia}                    &                        & 15                              & 0.66                          & 0.84                          & QA                                  \\
	\citet{jancso-etal-2020-acqdiv}               & ACQDIV                 & 14                              & 0.73\lowcov                   & 0.84                          & Language acquisition                \\
	\citet{vulic-etal-2020-multi}                 & Multi-SimLex           & 12                              & 0.58                          & 0.83                          & Semantic similarity                 \\
	\citet{eskander-etal-2020-morphagram}         & MorphAGram             & 12                              & 0.72\lowcov                   & 0.75                          & Morphological segmentation          \\
	\citet{hennig-etal-2023-multitacred}          & MultiTACRED            & 12                              & 0.58                          & 0.84                          & Relation extraction                 \\
	\citet{ponti-etal-2020-xcopa}                 & XCOPA                  & 12                              & 0.66                          & 0.75\missing                  & Commonsense reasoning               \\
	\citet{buechel-etal-2020-learning-evaluating} &                        & 12                              & 0.58                          & 0.72                          & Lexicon generation                  \\
	\citet{clark-etal-2020-tydi}                  & TyDiQA                 & 11                              & 0.67                          & 0.76                          & QA                                  \\
	\citet{zhang-etal-2021-mr}                    & Mr. TyDi               & 11                              & 0.67                          & 0.76                          & IR                                  \\
	\citet{shi2023language}                       & MGSM                   & 10                              & 0.65                          & 0.73                          & Arithmetic reasoning                \\
	\citet{mott-etal-2020-morphological}          &                        & 9                               & 0.65                          & 0.73                          & Morphological segmentation          \\
	\citet{srinivasan-choi-2022-tydip}            & TyDiP                  & 9                               & 0.62                          & 0.75                          & Politeness classification           \\
	\citet{gupta-etal-2023-cross}                 & Xtr-WikiQA \& TyDi-AS2 & 8                               & 0.64                          & 0.69                          & QA, Answer sentence selection       \\
	\citet{pfeiffer-etal-2022-xgqa}               & xGQA                   & 8                               & \cellcolor{low-color!40}0.58  & 0.66                          & Visual QA                           \\
	\citet{sproat-etal-2014-database}             &                        & 8                               & 0.58                          & 0.75                          & Linguistic information content      \\
	\citet{samir-silfverberg-2023-understanding}  &                        & 7                               & 0.66                          & 0.67                          & Morphological inflection            \\
	\citet{ginn-etal-2023-findings}               &                        & 7                               & 0.80\lowcov                   & 0.50                          & Interlinear glossing                \\
	\citet{sarti-etal-2022-divemt}                & DivEMT                 & 6                               & 0.64                          & 0.69                          & MT post-editing                     \\
	\citet{glavas-etal-2020-xhate}                & XHate-999              & 6                               & 0.64                          & 0.49                          & Abusive language classification     \\
	\citet{majewska-etal-2020-manual}             & SpAM                   & 5                               & 0.58                          & 0.70                          & Verb analysis                       \\
	\citet{yadavalli-etal-2023-slabert}           & SLABERT                & 5                               & 0.61                          & 0.66                          & Second language acquisition         \\
	\citet{liu-etal-2021-visually}                & MaRVL                  & 5                               & 0.69                          & 0.56                          & Visually grounded RE                \\
	\citet{hung-etal-2022-multi2woz}              & Multi2WOZ              & 4                               & 0.61                          & 0.58                          & Task-oriented dialogues             \\
	\citet{bartelds-etal-2023-making}             &                        & 4                               & --\lowcov                     & \cellcolor{low-color!40}0.00  & ASR                                 \\
	\citet{sasaki-etal-2002-co}                   &                        & 2                               & \cellcolor{high-color!40}0.84 & 0.39                          & Co-reference resolution             \\
	\bottomrule
\end{tabular}

    \caption{
        Papers that have a claim and introduce a dataset.
        \textbf{L} refers to the number of languages regarding the claim, with \nolangtext~indicating that one language does not have an ISO-639-3 code.
        \textbf{MPSD} is the Mean Pairwise Distance of syntactic URIEL features per language pair, with \lowcovtext~indicating that at least one language has low ($<5\%$) syntactic feature coverage.
        \textbf{FVI} refers to the Grambank feature value inclusion of the languages in the dataset: are all possible, non-missing values for a feature in Grambank included using the language selection.
        Here, \missingtext~refers to at least one language not being covered in Grambank, these languages are excluded from the coverage metric.
        The \textcolor{high-color}{highest} and \textcolor{low-color}{lowest} values for \textbf{MPSD} and \textbf{FVI} are highlighted.
        Abbreviations:
        \textit{QA}~=~Question Answering,
        \textit{RE}~=~Reasoning,
        \textit{NLI}~=~Natural Language Inference,
        \textit{IR}~=~Information Retrieval,
        \textit{ASR}~=~Automatic Speech Recognition,
        \textit{MT}~=~Machine Translation.
    }
    \label{tab:typ-datasets}
\end{table}

\newpage

\section{XTREME-R Results by Word Order}
\label{sec:app-word-order}
\noindent
In Table~\ref{tab:word-order} we can see that SVO is the most represented group for all tasks, with SOV coming in second.
After that, the representation of features tapers off quickly, with (the rather rare) OSV, OVS, and VOS word orders having no coverage at all.
We can see that the performance of models for SOV and VSO is consistently lower than SVO.
This can be due to the difference in the number of languages included for the group, although we did not see this effect for the inflectional feature (Table~\ref{tab:inflectional}).
\begin{table*}[h]
    \centering
    \tiny
    \begin{tabular}{ll||llr|llllllll}
	\toprule
	Subtask                                 & Model   & \textbf{Overall}    & \textbf{By F}       & $\Delta$      & OSV            & OVS            & VOS            & SVO                                            & SOV                                          & VSO                                          & NDO                                          & NA                                           \\\midrule\midrule
	\multirow{3}{*}{\textbf{MLQA}\f}        & XLM-R-L & 72.71 (\textit{7})  & 70.83 (\textit{7})  & \tgrad{-1.88} & - (\textit{0}) & - (\textit{0}) & - (\textit{0}) & \cellcolor{high-color!40}  75.22 (\textit{4})  & \cellcolor{low-color!40}  70.80 (\textit{1}) & \cellcolor{low-color!40}  67.00 (\textit{1}) & \cellcolor{low-color!40}  70.30 (\textit{1}) & - (\textit{0})                               \\
	                                        & mBERT   & 61.30 (\textit{7})  & 57.14 (\textit{7})  & \tgrad{-4.16} & - (\textit{0}) & - (\textit{0}) & - (\textit{0}) & \cellcolor{high-color!40}  66.85 (\textit{4})  & \cellcolor{low-color!40}  49.90 (\textit{1}) & \cellcolor{low-color!40}  51.60 (\textit{1}) & \cellcolor{low-color!40}  60.20 (\textit{1}) & - (\textit{0})                               \\
	                                        & mT5     & 75.59 (\textit{7})  & 74.06 (\textit{7})  & \tgrad{-1.53} & - (\textit{0}) & - (\textit{0}) & - (\textit{0}) & \cellcolor{high-color!40}  77.62 (\textit{4})  & \cellcolor{low-color!40}  75.30 (\textit{1}) & \cellcolor{low-color!40}  70.20 (\textit{1}) & \cellcolor{low-color!40}  73.10 (\textit{1}) & - (\textit{0})                               \\\midrule
	\multirow{2}{*}{\textbf{LAReQA}\map}    & XLM-R-L & 40.75 (\textit{11}) & 39.31 (\textit{11}) & \tgrad{-1.44} & - (\textit{0}) & - (\textit{0}) & - (\textit{0}) & \cellcolor{high-color!40}  42.10 (\textit{6})  & 39.75 (\textit{2})                           & \cellcolor{low-color!40}  34.60 (\textit{1}) & 40.80 (\textit{2})                           & - (\textit{0})                               \\
	                                        & mBERT   & 21.58 (\textit{11}) & 19.75 (\textit{11}) & \tgrad{-1.83} & - (\textit{0}) & - (\textit{0}) & - (\textit{0}) & \cellcolor{high-color!40}  24.10 (\textit{6})  & 15.10 (\textit{2})                           & \cellcolor{low-color!40}  17.00 (\textit{1}) & 22.80 (\textit{2})                           & - (\textit{0})                               \\\midrule
	\multirow{3}{*}{\textbf{TyDiQA}\f}      & XLM-R-L & 64.29 (\textit{9})  & 62.80 (\textit{9})  & \tgrad{-1.49} & - (\textit{0}) & - (\textit{0}) & - (\textit{0}) & \cellcolor{high-color!40}  67.26 (\textit{5})  & \cellcolor{low-color!40}  58.55 (\textit{2}) & \cellcolor{low-color!40}  62.60 (\textit{2}) & - (\textit{0})                               & - (\textit{0})                               \\
	                                        & mBERT   & 58.36 (\textit{9})  & 56.91 (\textit{9})  & \tgrad{-1.44} & - (\textit{0}) & - (\textit{0}) & - (\textit{0}) & \cellcolor{high-color!40}  61.24 (\textit{5})  & \cellcolor{low-color!40}  55.55 (\textit{2}) & \cellcolor{low-color!40}  53.95 (\textit{2}) & - (\textit{0})                               & - (\textit{0})                               \\
	                                        & mT5     & 81.94 (\textit{9})  & 81.45 (\textit{9})  & \tgrad{-0.50} & - (\textit{0}) & - (\textit{0}) & - (\textit{0}) & \cellcolor{high-color!40}  82.94 (\textit{5})  & \cellcolor{low-color!40}  78.40 (\textit{2}) & \cellcolor{low-color!40}  83.00 (\textit{2}) & - (\textit{0})                               & - (\textit{0})                               \\\midrule
	\multirow{3}{*}{\textbf{XQuAD}\f}       & XLM-R-L & 77.21 (\textit{11}) & 77.11 (\textit{11}) & \tgrad{-0.10} & - (\textit{0}) & - (\textit{0}) & - (\textit{0}) & \cellcolor{high-color!40}  76.70 (\textit{6})  & 76.55 (\textit{2})                           & \cellcolor{low-color!40}  74.40 (\textit{1}) & 80.80 (\textit{2})                           & - (\textit{0})                               \\
	                                        & mBERT   & 65.05 (\textit{11}) & 63.55 (\textit{11}) & \tgrad{-1.50} & - (\textit{0}) & - (\textit{0}) & - (\textit{0}) & \cellcolor{high-color!40}  67.35 (\textit{6})  & 56.60 (\textit{2})                           & \cellcolor{low-color!40}  62.20 (\textit{1}) & 68.05 (\textit{2})                           & - (\textit{0})                               \\
	                                        & mT5     & 81.54 (\textit{11}) & 81.04 (\textit{11}) & \tgrad{-0.49} & - (\textit{0}) & - (\textit{0}) & - (\textit{0}) & \cellcolor{high-color!40}  82.22 (\textit{6})  & 79.10 (\textit{2})                           & \cellcolor{low-color!40}  80.30 (\textit{1}) & 82.55 (\textit{2})                           & - (\textit{0})                               \\\midrule
	\multirow{3}{*}{\textbf{XNLI}\acc}      & XLM-R   & 79.24 (\textit{15}) & 78.57 (\textit{15}) & \tgrad{-0.67} & - (\textit{0}) & - (\textit{0}) & - (\textit{0}) & \cellcolor{high-color!40}  80.31 (\textit{9})  & 75.10 (\textit{3})                           & \cellcolor{low-color!40}  77.20 (\textit{1}) & 81.65 (\textit{2})                           & - (\textit{0})                               \\
	                                        & mBERT   & 66.51 (\textit{15}) & 65.79 (\textit{15}) & \tgrad{-0.72} & - (\textit{0}) & - (\textit{0}) & - (\textit{0}) & \cellcolor{high-color!40}  68.19 (\textit{9})  & 60.03 (\textit{3})                           & \cellcolor{low-color!40}  66.00 (\textit{1}) & 68.95 (\textit{2})                           & - (\textit{0})                               \\
	                                        & mT5     & 84.85 (\textit{15}) & 84.71 (\textit{15}) & \tgrad{-0.14} & - (\textit{0}) & - (\textit{0}) & - (\textit{0}) & \cellcolor{high-color!40}  85.39 (\textit{9})  & 81.83 (\textit{3})                           & \cellcolor{low-color!40}  84.50 (\textit{1}) & 87.10 (\textit{2})                           & - (\textit{0})                               \\\midrule
	\multirow{2}{*}{\textbf{Mewsli-X}\map}  & XLM-R-L & 45.75 (\textit{11}) & 45.66 (\textit{11}) & \tgrad{-0.09} & - (\textit{0}) & - (\textit{0}) & - (\textit{0}) & \cellcolor{high-color!40}  53.16 (\textit{5})  & 35.98 (\textit{4})                           & \cellcolor{low-color!40}  28.70 (\textit{1}) & \cellcolor{low-color!40}  64.80 (\textit{1}) & - (\textit{0})                               \\
	                                        & mBERT   & 38.58 (\textit{11}) & 37.88 (\textit{11}) & \tgrad{-0.71} & - (\textit{0}) & - (\textit{0}) & - (\textit{0}) & \cellcolor{high-color!40}  47.28 (\textit{5})  & 27.93 (\textit{4})                           & \cellcolor{low-color!40}  15.30 (\textit{1}) & \cellcolor{low-color!40}  61.00 (\textit{1}) & - (\textit{0})                               \\
	\midrule
	\midrule
	\multirow{2}{*}{\textbf{Tatoeba}\acc}   & XLM-R   & 77.29 (\textit{41}) & 72.82 (\textit{38}) & \tgrad{-4.46} & - (\textit{0}) & - (\textit{0}) & - (\textit{0}) & \cellcolor{high-color!40}  81.42 (\textit{18}) & 75.74 (\textit{14})                          & \cellcolor{low-color!40}  64.55 (\textit{2}) & 86.57 (\textit{4})                           & 55.83 (\textit{3})                           \\
	                                        & mBERT   & 43.33 (\textit{41}) & 39.66 (\textit{38}) & \tgrad{-3.67} & - (\textit{0}) & - (\textit{0}) & - (\textit{0}) & \cellcolor{high-color!40}  52.57 (\textit{18}) & 33.04 (\textit{14})                          & \cellcolor{low-color!40}  25.10 (\textit{2}) & 54.78 (\textit{4})                           & 32.83 (\textit{3})                           \\\midrule
	\multirow{3}{*}{\textbf{XCOPA}\acc}     & XLM-R   & 69.22 (\textit{11}) & 66.16 (\textit{9})  & \tgrad{-3.06} & - (\textit{0}) & - (\textit{0}) & - (\textit{0}) & \cellcolor{high-color!40}  72.89 (\textit{7})  & \cellcolor{low-color!40}  72.90 (\textit{2}) & - (\textit{0})                               & - (\textit{0})                               & \cellcolor{low-color!40}  52.70 (\textit{2}) \\
	                                        & mBERT   & 56.05 (\textit{11}) & 55.17 (\textit{9})  & \tgrad{-0.88} & - (\textit{0}) & - (\textit{0}) & - (\textit{0}) & \cellcolor{high-color!40}  57.11 (\textit{7})  & \cellcolor{low-color!40}  55.50 (\textit{2}) & - (\textit{0})                               & - (\textit{0})                               & \cellcolor{low-color!40}  52.90 (\textit{2}) \\
	                                        & mT5     & 74.89 (\textit{11}) & 72.62 (\textit{9})  & \tgrad{-2.27} & - (\textit{0}) & - (\textit{0}) & - (\textit{0}) & \cellcolor{high-color!40}  77.61 (\textit{7})  & \cellcolor{low-color!40}  77.00 (\textit{2}) & - (\textit{0})                               & - (\textit{0})                               & \cellcolor{low-color!40}  63.25 (\textit{2}) \\\midrule
	\multirow{2}{*}{\textbf{WikiANN-NER}\f} & XLM-R-L & 64.43 (\textit{48}) & 65.10 (\textit{43}) & \tgrad{+0.67} & - (\textit{0}) & - (\textit{0}) & - (\textit{0}) & \cellcolor{high-color!40}  66.80 (\textit{20}) & 59.74 (\textit{17})                          & \cellcolor{low-color!40}  57.95 (\textit{2}) & 79.70 (\textit{4})                           & 61.30 (\textit{5})                           \\
	                                        & mBERT   & 62.68 (\textit{48}) & 63.98 (\textit{43}) & \tgrad{+1.30} & - (\textit{0}) & - (\textit{0}) & - (\textit{0}) & \cellcolor{high-color!40}  67.34 (\textit{20}) & 54.26 (\textit{17})                          & \cellcolor{low-color!40}  58.80 (\textit{2}) & 75.92 (\textit{4})                           & 63.58 (\textit{5})                           \\\midrule
	\multirow{2}{*}{\textbf{UD-POS}\f}      & XLM-R-L & 74.96 (\textit{38}) & 77.45 (\textit{36}) & \tgrad{+2.50} & - (\textit{0}) & - (\textit{0}) & - (\textit{0}) & \cellcolor{high-color!40}  74.59 (\textit{20}) & 69.65 (\textit{10})                          & \cellcolor{low-color!40}  71.55 (\textit{2}) & 86.97 (\textit{4})                           & \cellcolor{low-color!40}  84.50 (\textit{2}) \\
	                                        & mBERT   & 70.90 (\textit{38}) & 71.60 (\textit{36}) & \tgrad{+0.69} & - (\textit{0}) & - (\textit{0}) & - (\textit{0}) & \cellcolor{high-color!40}  72.74 (\textit{20}) & 62.82 (\textit{10})                          & \cellcolor{low-color!40}  61.25 (\textit{2}) & 83.22 (\textit{4})                           & \cellcolor{low-color!40}  77.95 (\textit{2}) \\
	\bottomrule
\end{tabular}

    \caption{XTREME-R task and model results, grouped by word order (WALS 81A). \textbf{Overall} refers to the average over all languages. 
    \textbf{By~F}eature is the average of the WALS feature averages, excluding languages for which there is no coverage. 
    The delta shows the difference of \textbf{Overall} and \textbf{By F}. 
    Tasks that have coverage for all their included languages are on top, those partially covered are in the bottom portion. The word orders are listed in the following columns, with \emph{NDO} being 
    \emph{No Dominant Order} and \emph{NA} referring to \emph{Not Available} (in WALS).
    The \textcolor{high-color}{highest} and the \textcolor{low-color}{lowest} number of languages per grouping are highlighted. The \emph{(italicized)} number refers to the number of languages in a particular subset. Metrics: \acctext~=~Accuracy, \maptext~=~mAP@20, \ftext~=~F1.}
    \label{tab:word-order}
\end{table*}

\newpage
\section{XTREME-R Grambank Feature Value Inclusion}
\label{app:pca}
\begin{figure*}[!h]
    \captionsetup[sub]{font=scriptsize,belowskip=0.2em}
     \centering
     \begin{subfigure}[b]{0.30\textwidth}
        \centering
        \includegraphics[width=\textwidth]{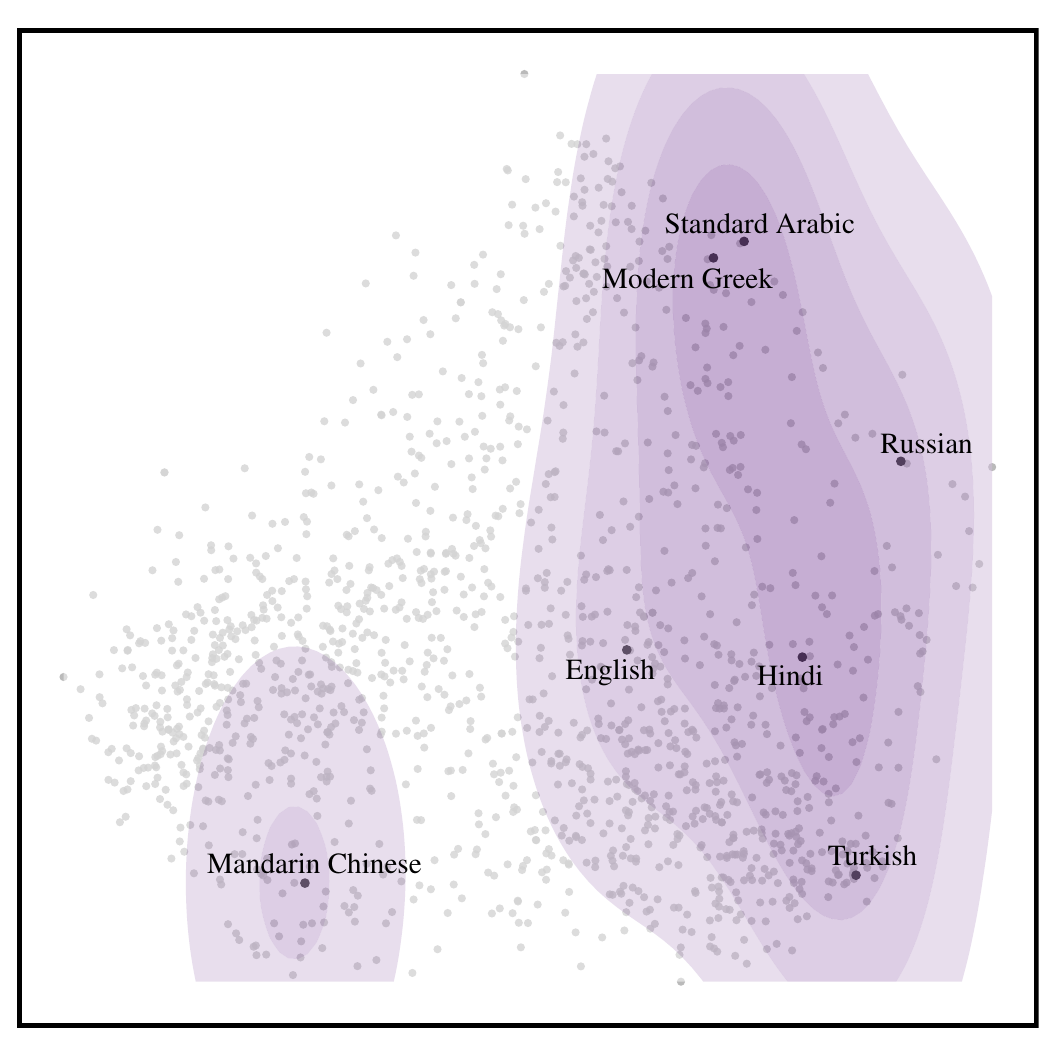}
        \caption{LAReQA \cite{roy-etal-2020-lareqa}.}
        \label{fig:xtremer-lareqa}
     \end{subfigure}
     \begin{subfigure}[b]{0.30\textwidth}
        \centering
        \includegraphics[width=\textwidth]{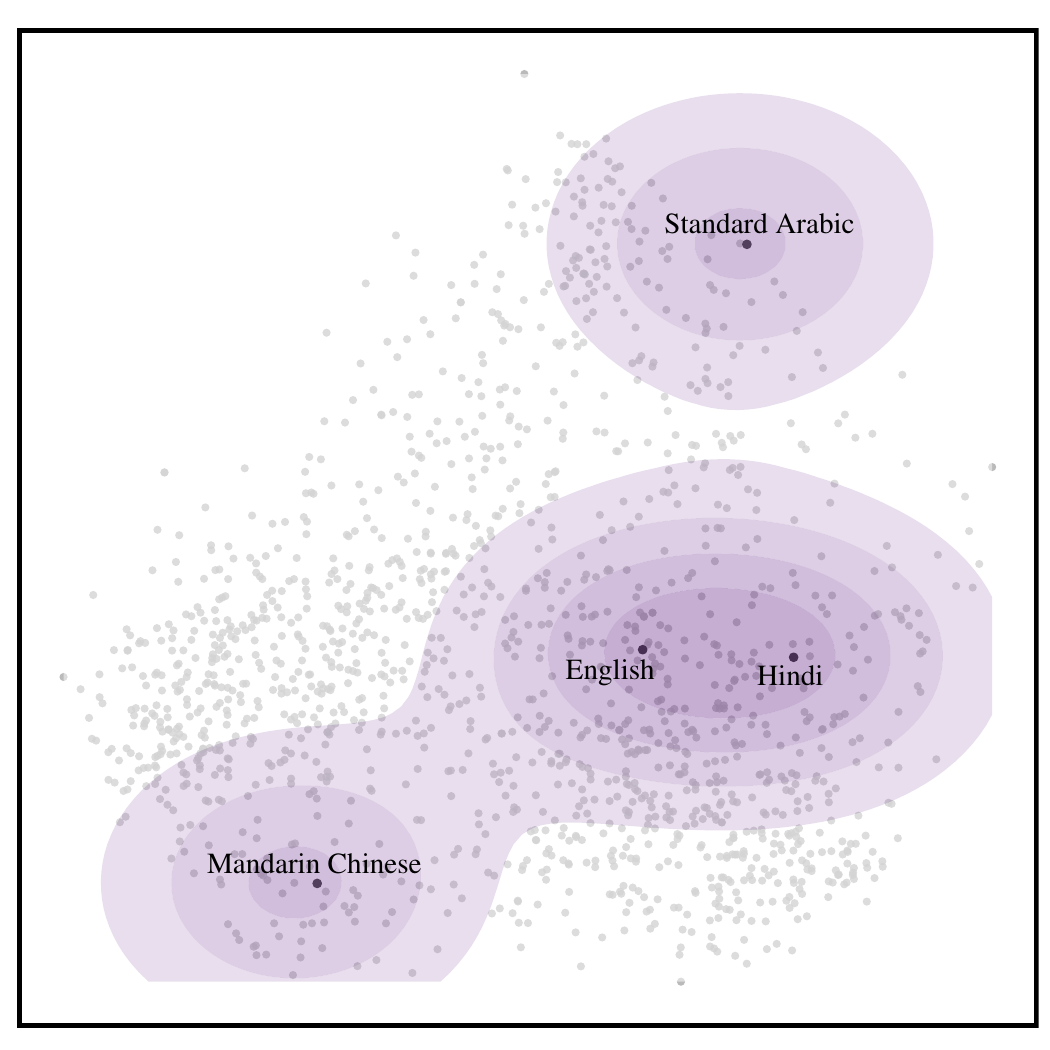}
        \caption{MLQA \cite{lewis-etal-2020-mlqa}.}
        \label{fig:xtremer-mlqa}
     \end{subfigure}
     \begin{subfigure}[b]{0.30\textwidth}
        \centering
        \includegraphics[width=\textwidth]{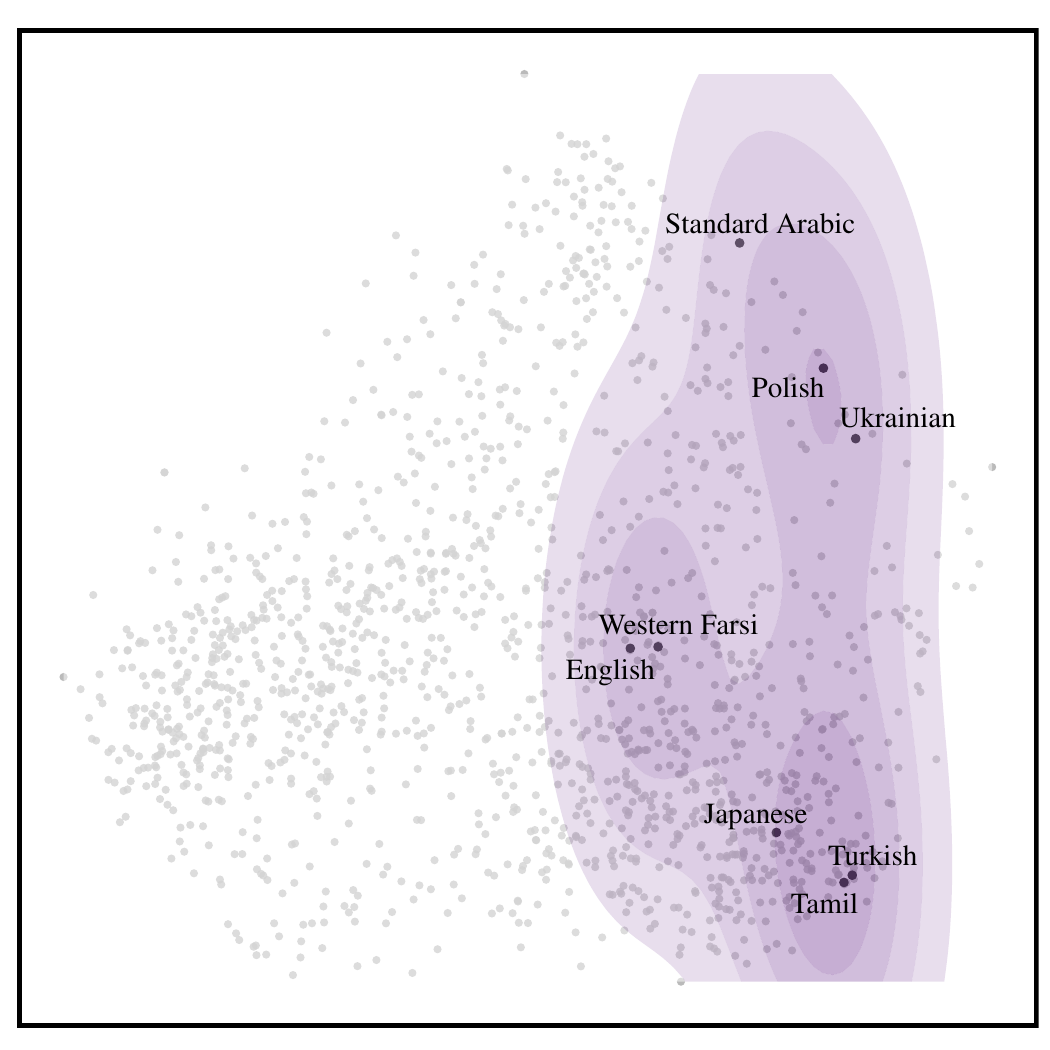}
        \caption{Mewsli-X \cite{botha-etal-2020-entity}.}
        \label{fig:xtremer-mewsli-x}
     \end{subfigure}
     
     \begin{subfigure}[b]{0.30\textwidth}
        \centering
        \includegraphics[width=\textwidth]{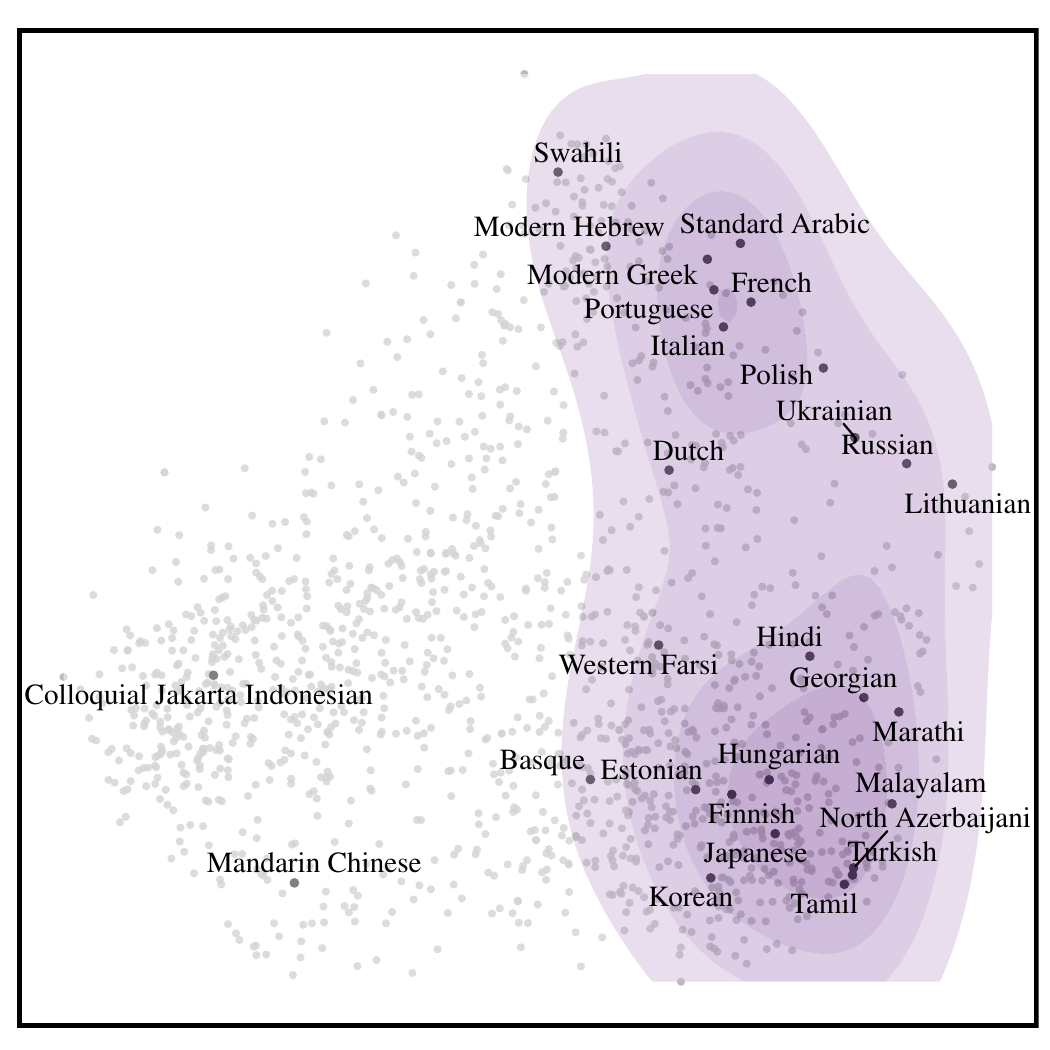}
        \caption{Tatoeba \cite{artetxe-schwenk-2019-massively}.} %
        \label{fig:xtremer-tatoeba}
     \end{subfigure}
     \begin{subfigure}[b]{0.30\textwidth}
        \centering
        \includegraphics[width=\textwidth]{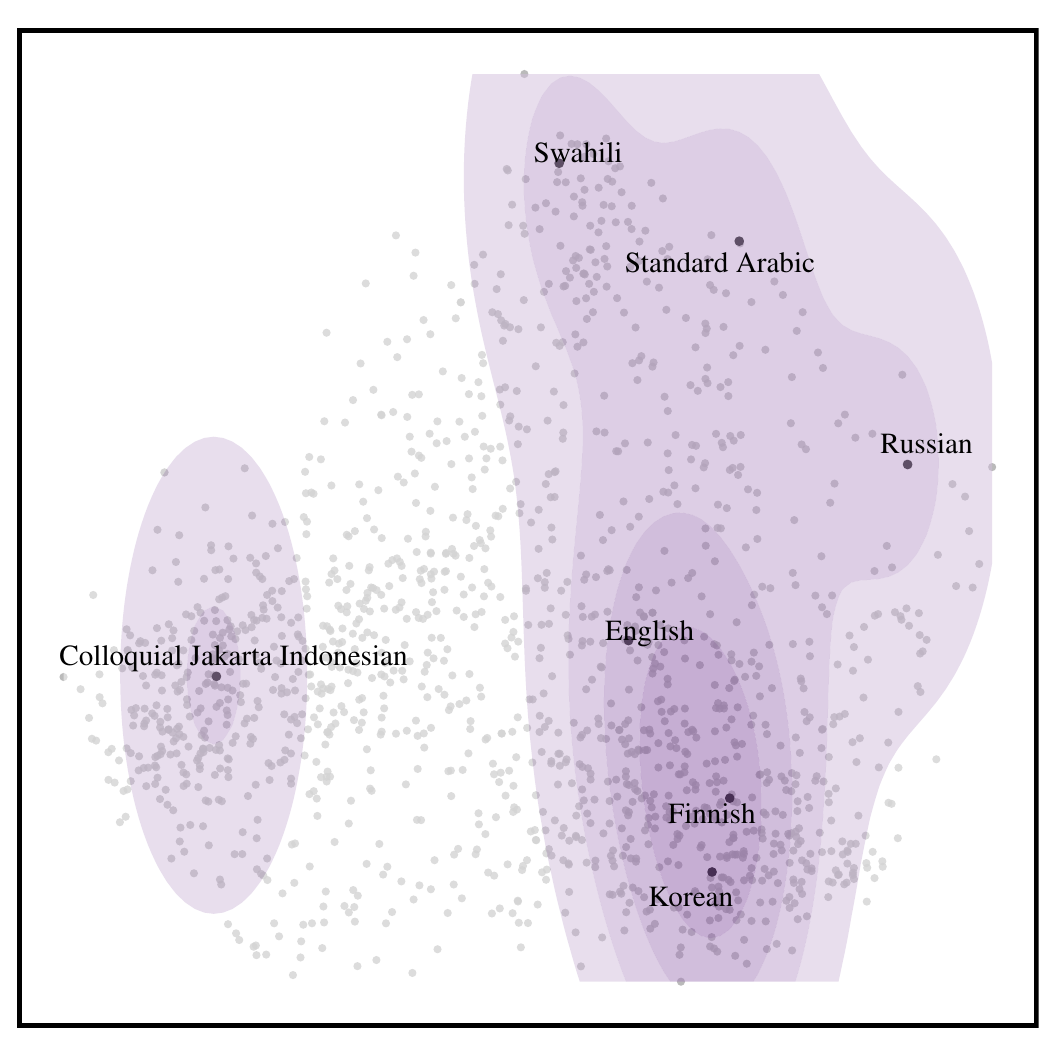}
        \caption{TyDiQA \cite{clark-etal-2020-tydi}.}
        \label{fig:xtremer-tydiqa}
     \end{subfigure}
     \begin{subfigure}[b]{0.30\textwidth}
        \centering
        \includegraphics[width=\textwidth]{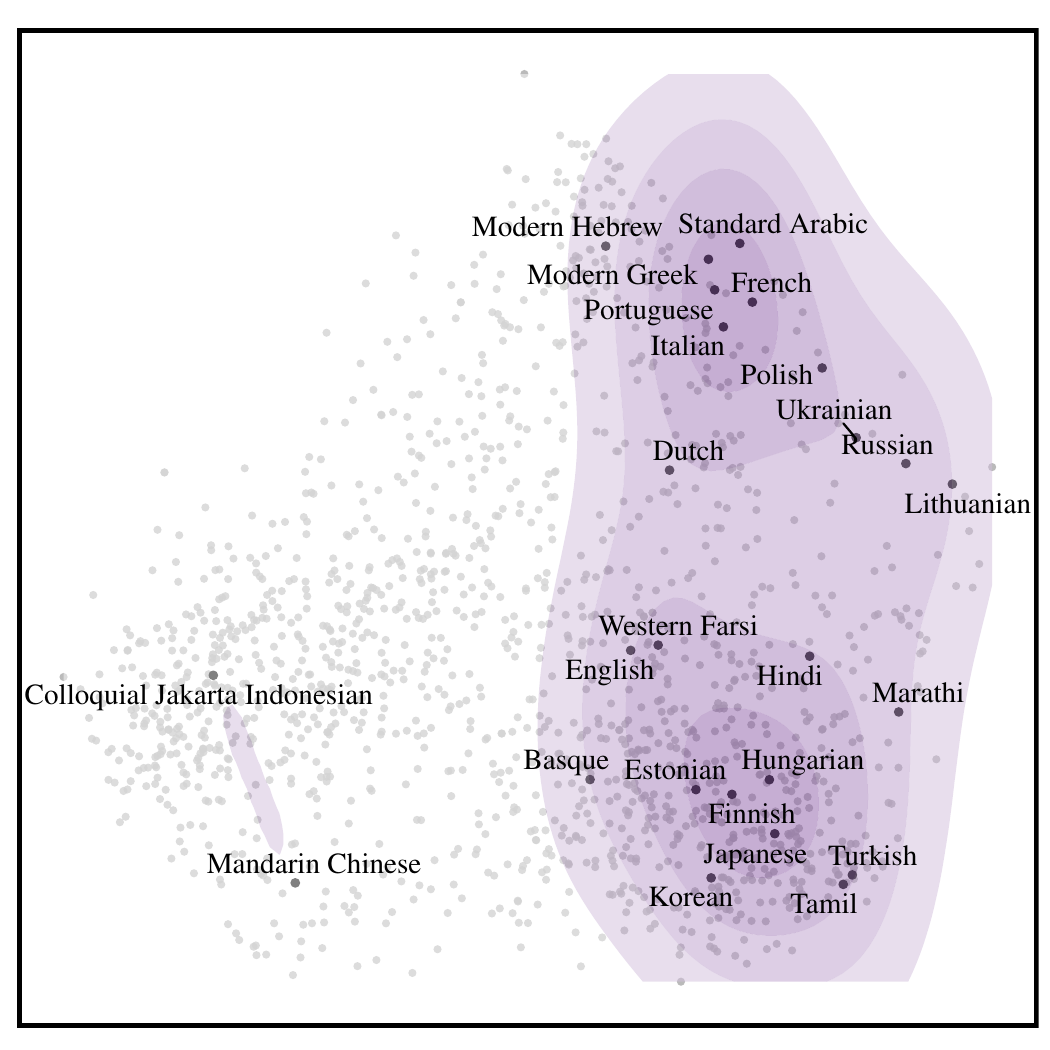}
        \caption{UD-POS \cite{nivre2018universal}.}
        \label{fig:xtremer-ud-pos}
     \end{subfigure}

      \begin{subfigure}[b]{0.30\textwidth}
        \centering
        \includegraphics[width=\textwidth]{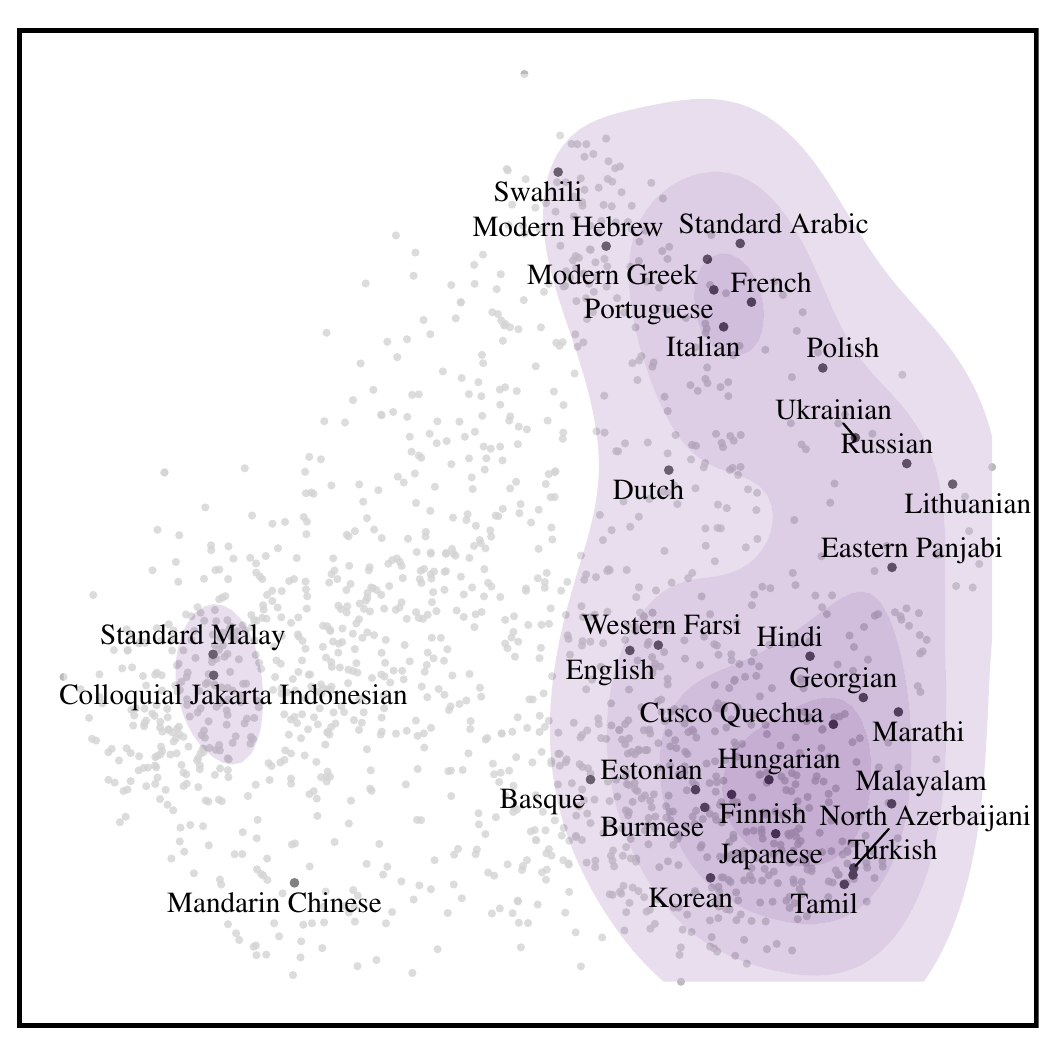}
        \caption{WikiANN-NER \cite{pan-etal-2017-cross}.}
        \label{fig:xtremer-wikiann-ner}
     \end{subfigure}
     \begin{subfigure}[b]{0.30\textwidth}
        \centering
        \includegraphics[width=\textwidth]{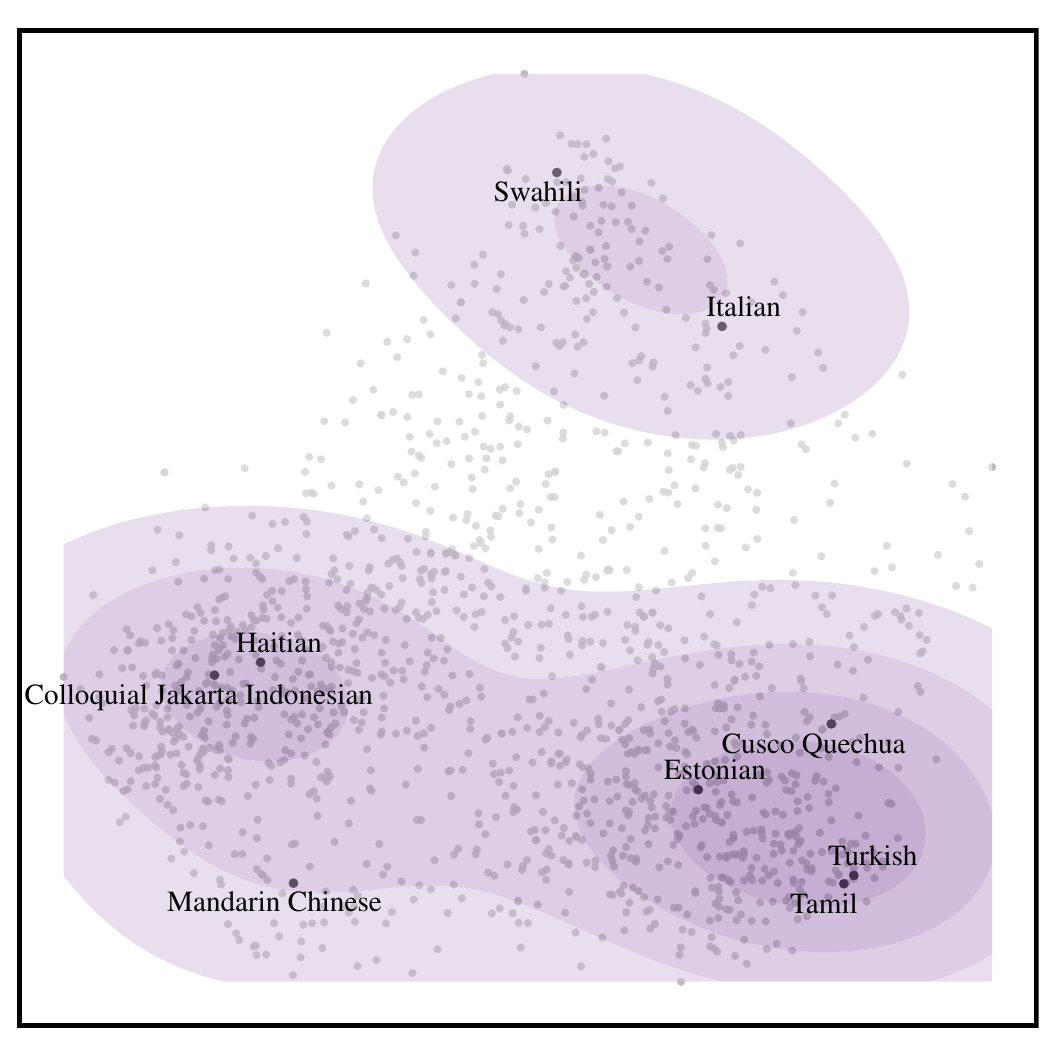}
        \caption{XCOPA \cite{ponti-etal-2020-xcopa}.}
        \label{fig:xtremer-xcopa}
     \end{subfigure}
     \begin{subfigure}[b]{0.30\textwidth}
        \centering
        \includegraphics[width=\textwidth]{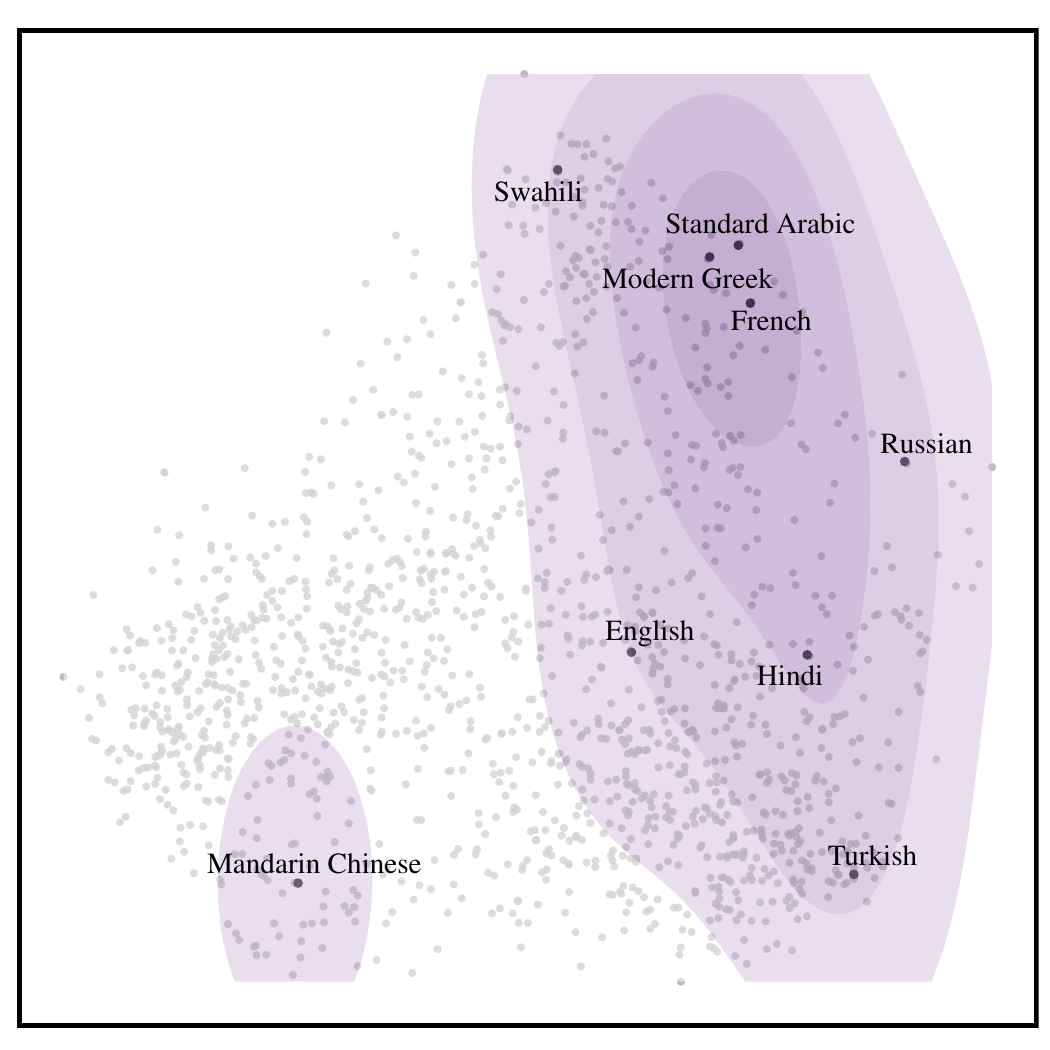}
        \caption{XNLI \cite{conneau-etal-2018-xnli}.}
        \label{fig:xtremer-xnli}
     \end{subfigure}

     \begin{subfigure}[b]{0.30\textwidth}
        \centering
        \includegraphics[width=\textwidth]{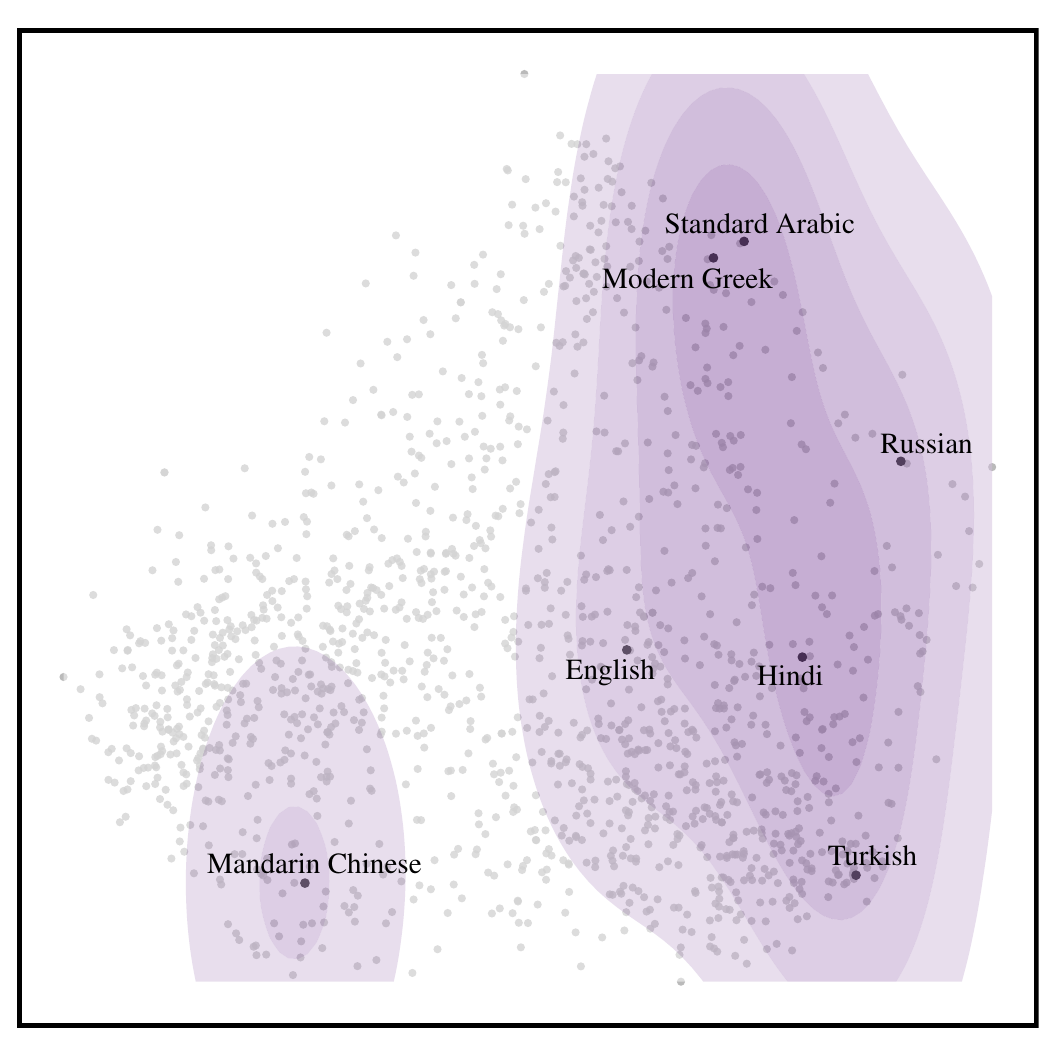}
        \caption{XQuAD \cite{artetxe-etal-2020-cross}.}
        \label{fig:xtremer-xquad}
     \end{subfigure}
     \begin{subfigure}[b]{0.30\textwidth}
        \centering
        \includegraphics[width=\textwidth]{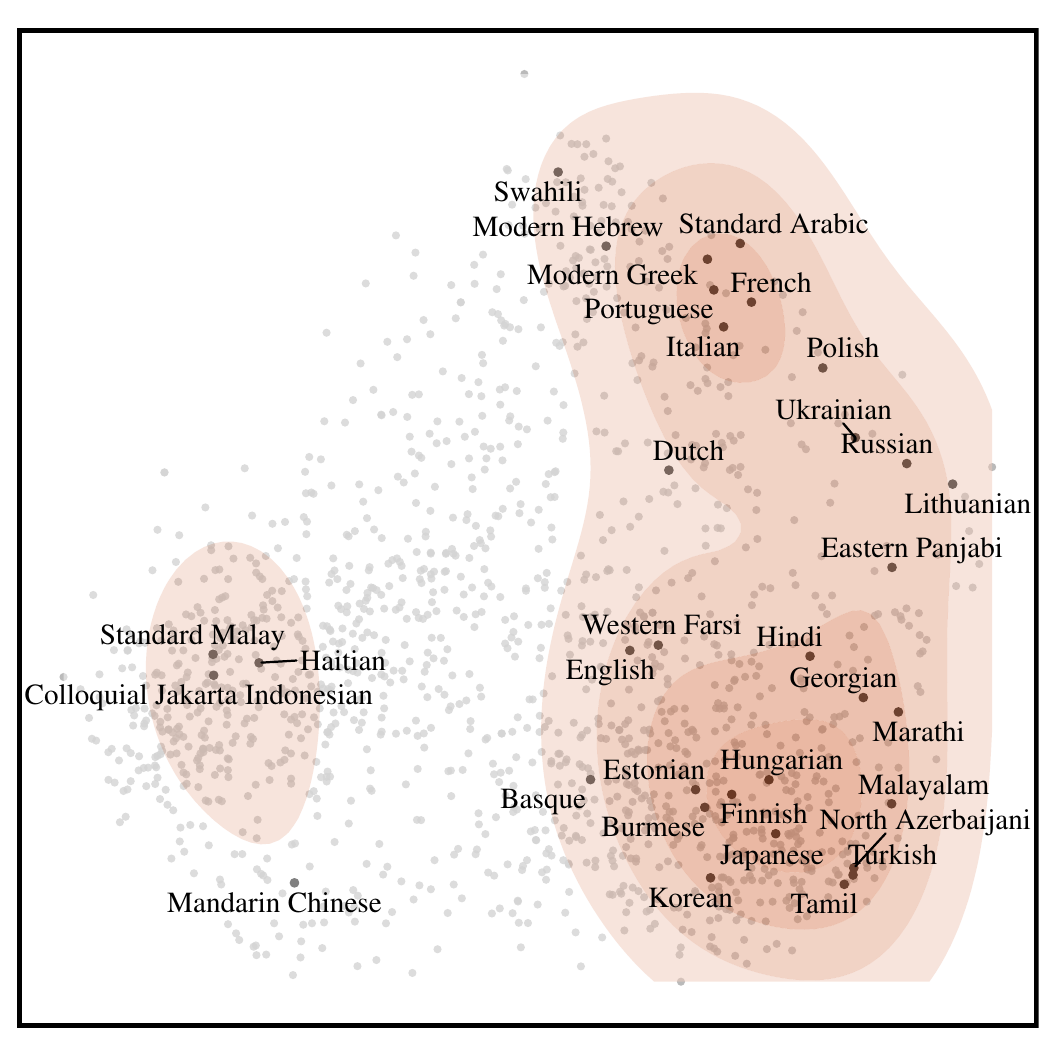}
        \caption{All subtasks together.}
        \label{fig:xtremer-combined}
     \end{subfigure}
     
    \caption{PCA Grambank feature value inclusion per XTREME-R subtask. Where applicable, we cite the original dataset that XTREME-R includes or used as a starting point to expand upon.}
    \label{fig:xtremer-pca}
\end{figure*}

\end{document}